\documentclass{article} 
\usepackage[preprint]{colm2026_conference}
\usepackage{etoc}
\usepackage[most]{tcolorbox}
\usepackage{microtype}
\usepackage{hyperref}
\usepackage{url}
\usepackage{booktabs}
\usepackage{amsmath}
\usepackage{amsthm}

\newtheorem{remark}{Remark}
\usepackage{booktabs}   
\usepackage{multirow}   
\usepackage{graphicx}   
\usepackage{subcaption} 
\usepackage[ruled,linesnumbered]{algorithm2e}
\usepackage[most]{tcolorbox}
\usepackage{minted}
\usepackage{wrapfig}
\usepackage{xcolor}
\usepackage{colortbl}
\usepackage{pifont}
\usepackage{algorithmic}
\newcommand{\cmark}{\ding{51}}

\definecolor{richpurple}{RGB}{75,46,131}
\definecolor{beige}{RGB}{245,245,220}

\newcommand{\purplecomic}[1]{%
  {\color{richpurple}\selectfont #1}%
}
\definecolor{groupcolor}{RGB}{240, 240, 240}
\definecolor{mygreen}{RGB}{34, 139, 34}
\definecolor{myred}{RGB}{220, 20, 60}
\definecolor{teachercolor}{RGB}{255, 243, 205}
\definecolor{ourscolor}{RGB}{219, 234, 254}
\definecolor{mygreen}{RGB}{34, 139, 34}
\definecolor{myred}{RGB}{220, 20, 60}
\definecolor{b2fcolor}{RGB}{235, 230, 245}   
\definecolor{f2bcolor}{RGB}{245, 230, 238}   

\definecolor{boxback}{RGB}{242, 240, 248}    
\definecolor{boxframe}{RGB}{112, 108, 145}  
\newtcolorbox{findingbox}[1]{
  colback=boxback,        
  colframe=boxframe,       
  fonttitle=\bfseries,    
  title={#1},             
  coltitle=white,         
  sharp corners=south,    
  rounded corners=all,    
  arc=4pt,                
  boxrule=0.5pt,          
  top=2pt,                
  bottom=2pt,             
  left=5pt,               
  right=5pt               
}

\usepackage{lineno}
\newcommand{\ours}{TCOD\xspace}
\newcommand{\oursfull}{Temporal Curriculum On-Policy Distillation\xspace}
\definecolor{darkblue}{rgb}{0, 0, 0.5}
\hypersetup{colorlinks=true, citecolor=darkblue, linkcolor=darkblue, urlcolor=darkblue}

\etocdepthtag.toc{mtchapter}

\title{\purplecomic{\textbf{TCOD}}: Exploring Temporal Curriculum in On-Policy Distillation for Multi-turn Autonomous Agents}

\author{
Jiaqi Wang\footnotemark[2],
~Wenhao Zhang, 
~Weijie Shi, 
~Yaliang Li, 
~James Cheng\footnotemark[2] \\
[1em]
Tongyi Lab\includegraphics[height=12pt]{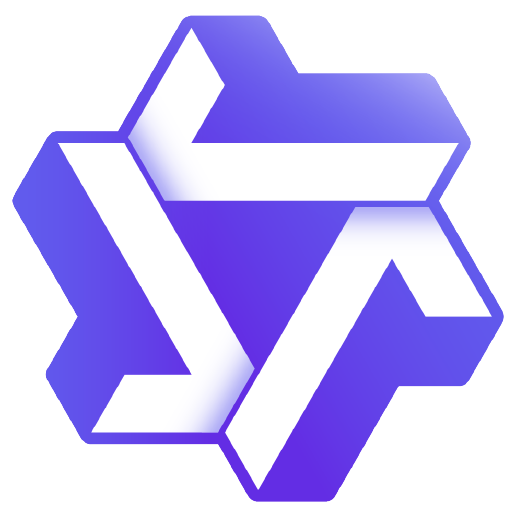}, Alibaba Group \\
[1em]
}

\begin{document}

\ifcolmsubmission
\linenumbers
\fi

\maketitle

\begingroup
  \renewcommand\thefootnote{}%
  \footnotetext{
      $^{\dagger}$ The Chinese University of Hong Kong.
  }
\endgroup
\phantomsection

\begin{tcolorbox}[
  colback=blue!5!white,
  colframe=richpurple!80!black,
  boxrule=1.5pt,
  arc=3mm,
  left=4mm,
  right=4mm,
  top=3mm,
  bottom=3mm,
  fonttitle=\bfseries,
  title=\textbf{Abstract}
]

On-policy distillation (OPD) has shown strong potential for transferring reasoning ability from frontier or domain-specific models to smaller students. 
While effective on static single-turn tasks, its behavior in multi-turn agent settings remains underexplored. 
In this work, we identify a key limitation of vanilla OPD in such settings, which we term \textit{Trajectory-Level KL Instability}.
Specifically, we observe that KL divergence increases together with a drop in success rate, and even after convergence, the KL remains high, leading to unstable training.
This instability arises from inter-turn error compounding: as errors accumulate, the student is driven beyond the teacher’s effective support, rendering the supervision signal unreliable.
To address this, we propose \textbf{TCOD} (\textbf{T}emporal \textbf{C}urriculum \textbf{O}n-Policy \textbf{D}istillation), a simple yet effective framework that controls the trajectory depth exposed to the student and progressively expands it from short to long with a curriculum schedule. 
Experimental results across four student–teacher pairs on three multi-turn agent benchmarks (ALFWorld, WebShop, ScienceWorld) show that \textit{\ours mitigates KL escalation and enhances KL stability throughout training, improving agent performance by up to 18 points over vanilla OPD.}
Further evaluations show that \ours can even \textit{surpass the teacher’s performance and generalize to tasks on which the teacher fails.}

\end{tcolorbox}

\section{Introduction}

On-policy distillation has recently emerged as a primary paradigm for transferring complex reasoning capabilities from frontier models to their smaller counterparts~\citep{agarwal2024policy,lu2025onpolicydistillation}. 
These methods have demonstrated remarkable success in mathematical or question answering tasks by minimizing token-level KL divergence over student-generated rollouts~\citep{jang2026stable, ko2026scaling,jin2026entropy}. 
However, these approaches are inherently designed for static, single-turn reasoning.
Consequently, they leave the more challenging \textit{multi-turn} agent setting underexplored, where the model must continuously reason and act based on a growing history of sequential interactions.
It remains a critical open question whether the stability of vanilla OPD can safely generalize to such dynamic, long-horizon environments.

\begin{figure}[t]
    \centering
    \includegraphics[width=0.99\linewidth]{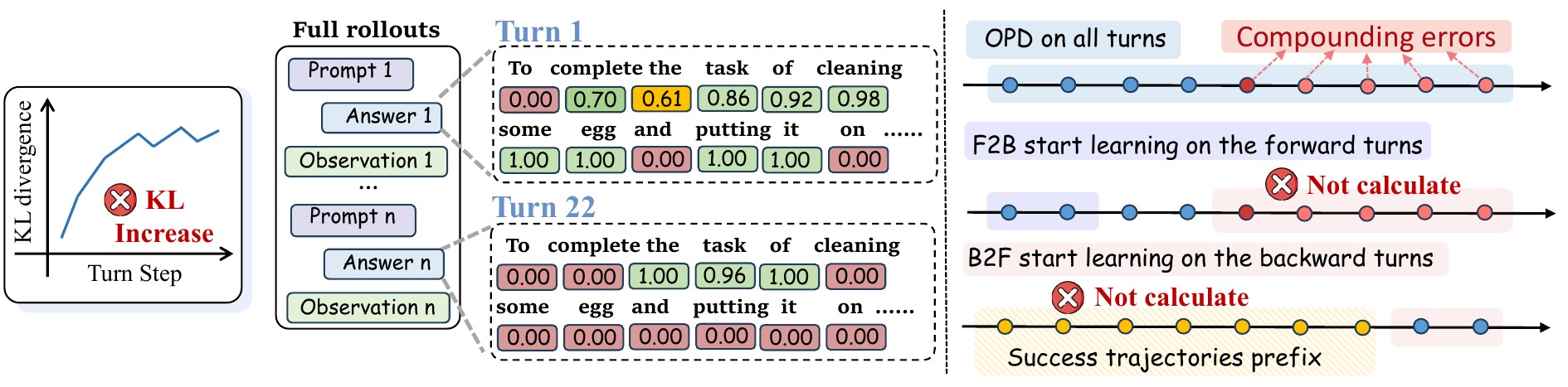}
    \vspace{-0.5em}
    \caption{ \textbf{(left)} In OPD for multi-turn agents, as the number of turns increases, the teacher assigns progressively lower probabilities to tokens in student-generated responses, indicating increasing KL divergence at each turn, rendering the supervision signal unreliable. \textbf{(right)} OPD uses all turns and thus includes compounding errors, whereas \ours-F2B/B2F progressively expands from short to long trajectories, alleviating calculating the error turns.}
    \label{fig:intro}
    \vspace{-1em}
\end{figure}

In this work, we present empirical evidence that naively applying vanilla OPD in this multi-turn regime leads to a fundamental failure mode, which we term \textbf{Trajectory-Level KL Instability}.
Through experiments on ALFWorld~\citep{shridhar2020alfworld}, we find that \textbf{\textit{(i)}} the student models suffer from simultaneous KL escalation and success rate collapse, 
and \textbf{\textit{(ii)}} although they eventually converge, they begin with prohibitively high KL divergence, both of which induce training instability. 
Crucially, as shown in Figure~\ref{fig:intro}(left), we reveal the underlying mechanism: 
Compounding errors across turns progressively push the student into states outside the teacher’s effective support. 
As a result, the teacher assigns lower probabilities to tokens in student-generated responses, indicating increasing KL divergence at each turn and rendering its supervision signal unreliable.

To address this, we propose \textbf{\ours} (\textbf{T}emporal \textbf{C}urriculum \textbf{O}n-Policy \textbf{D}istillation), a simple yet effective framework that controls the trajectory depth exposed to the student and progressively expands it from short to long via a pacing strategy governed by a configurable curriculum growth rate.
Based on this core idea, as shown in Figure~\ref{fig:intro}(right), we introduce two practical variants with only minimal code modifications: \textbf{Forward-to-Backward (\ours-F2B)}, which restricts the student to the early steps of the trajectory and progressively extends it to the maximum explore horizon; 
and \textbf{Backward-to-Forward (\ours-B2F)}, which leverages the teacher to navigate the agent to near-terminal states, alleviating error accumulation in the early steps, while gradually extending the student's rollout horizon backward to the initial stages.

Built on top of \ours-F2B/B2F, we evaluate four student-teacher pairs on three multi-turn agent benchmarks: ALFWorld~\citep{shridhar2020alfworld}, 
WebShop~\citep{yao2022webshop}, and ScienceWorld~\citep{wang2022scienceworld}. 
Overall, \textbf{\ours alleviates KL instability and improves performance} by recovering Qwen3-1.7B from near-zero success rates and boosting the larger one (e.g., Qwen2.5-7B) by up to $15.71$ success rate points while reducing action rounds by an average of $2.97$ steps. 
Moreover, \ours does not merely imitate the teacher---on the hard split of ALFWorld where the teacher fails under pass@10 sampling, \ours-B2F surpasses 
the teacher's success rate by up to $14$ points, \textbf{demonstrating generalization beyond the teacher's own capability boundary.} 
Finally, \ours-F2B/B2F are robust to the curriculum growth rate with less than $2\%$ performance variation, and reduce total training time by up to $32\%$ compared to vanilla OPD.

\section{Related Work}


%

\textbf{LLM-based Multi-turn Agents.}
Large language models have demonstrated strong capabilities as multi-turn agents~\citep{openai2024gpt4technicalreport,yang2025qwen3technicalreport}. 
A common paradigm interleaves reasoning and action generation via frameworks such as ReAct~\citep{yao2022react}, enabling agents to solve tasks in embodied planning, web navigation, and other interactive environments~\citep{shridhar2020alfworld,yao2022webshop,wang2022scienceworld,merrill2026terminal,cli_anything,li2026skillsbench}.
Recent systems such as OpenClaw~\citep{wang2026openclaw,openclaw} further demonstrate the potential of LLM-based agents for long-horizon tasks, motivating increasing interest in general-purpose agentic frameworks.
Despite these advances, training multi-turn agents remains challenging due to long-horizon credit assignment~\citep{Guo_2025}, memory management~\citep{shi2026r}, and the sample inefficiency of reinforcement learning in sparse-reward settings~\citep{feng2025groupingrouppolicyoptimizationllm,penaloza2026privileged}.

\textbf{On-Policy Distillation and its Limitations.}
On-Policy Distillation~\citep{agarwal2024policy, lu2025onpolicydistillation} has emerged as a compelling alternative to on-policy post-training by replacing sparse scalar rewards with dense distillation signals and thereby improving sample efficiency.
Existing work has improved OPD through several design choices, including objective design~\citep{jang2026stable, jin2026entropy}, optimization heuristics~\citep{ko2026scaling}, and alternative supervision sources~\citep{ye2026policy, zhao2026self}. 
These methods, such as balancing forward and backward KL terms~\citep{jang2026stable, jin2026entropy} and incorporating RL-style heuristics such as reward clipping~\citep{ko2026scaling}, improve training stability and convergence.
However, these approaches are primarily designed for single-turn settings and do not directly address multi-turn agent environments.

\textbf{Curriculum Learning.}
Curriculum learning~\citep{bengio2009curriculum} is a training strategy where a model is exposed to progressively more difficult examples as its competence grows. 
Recent works~\citep{zhang2026beyond, wang2025dump} apply this to the pre-training and post-training of LLMs, respectively. 
\citet{shi2025efficient, wang2025practitioner,gong2026temp} further apply curriculum learning to reinforcement learning methods, such as GRPO~\cite{Guo_2025}, but still rely on an external model to measure difficulty.
\citet{lauffer2025imitation} trains the student only on the expert's subsequent corrective actions, breaking the on-policy setting.
Our approach avoids both by defining difficulty through increasing trajectory depth, using only student-generated data, keeping training simple, on-policy, and more stable.


\section{Preliminary}

In this paper, we consider multi-turn autonomous agents interacting with an 
environment over a finite horizon. Let $t \in \{0, \dots, T-1\}$ 
denote the turn index within a trajectory, where $T$ is the maximum 
number of interaction steps. At each turn $t$, the agent receives an 
observation $o_t$, generates a response $a_t$, and the environment 
returns the next observation $o_{t+1}$. 
Following the recent agent frameworks~\citep{wang2025thinknotselectivereasoning}, each response $a_t$ consists of a chain-of-thought 
reasoning trace followed by an executable action.

\textbf{History State for Multi-turn Agent.} Since the environment is generally partially observable, we define the 
agent state as the full interaction history up to the current 
observation:
\begin{equation}
    h_t = (o_0, a_0, o_1, a_1, \dots, o_{t-1}, a_{t-1}, o_t).
\end{equation}
A complete trajectory is then $\tau = (h_0, a_0, h_1, a_1, \dots, 
h_{T-1}, a_{T-1})$, which terminates either when a termination action 
is taken or when the horizon $T$ is reached.

\textbf{On-Policy Distillation for Multi-turn Agent.}
Given a teacher policy $\pi_\phi$ and a student policy $\pi_\theta$, 
the goal of on-policy distillation is to align the student with the 
teacher under the student's own state distribution. The objective is:
\begin{equation}
    \mathcal{L}_{\text{OPD}}(\theta)
    = \mathbb{E}_{\tau \sim \pi_\theta}
    \left[
    \sum_{t=0}^{T-1}
    \mathcal{D}_{\mathrm{KL}}
    \big(
    \pi_\phi(a_t \mid h_t)
    \parallel
    \pi_\theta(a_t \mid h_t)
    \big)
    \right],
\end{equation}
where $\mathcal{D}_{\mathrm{KL}}(\pi_\phi \parallel \pi_\theta) = \sum_{a_t} \pi_\phi(a_t \mid h_t) \log \frac{\pi_\phi(a_t \mid h_t)}{\pi_\theta(a_t \mid h_t)}$ is the KL divergence measuring the discrepancy between the teacher policy $\pi_\phi$ and the student policy $\pi_\theta$.

\section{\ours: Temporal Curriculum On-Policy Distillation}
In this section, we observe a key limitation of OPD in multi-turn agent settings, termed trajectory-level KL instability. 
Through empirical analysis, we show that OPD exhibits instability in long-horizon interactions, where compounding errors lead to escalating KL divergence and degraded performance.
Motivated by these findings, we propose \ours, a temporal curriculum strategy that progressively controls trajectory depth during training to improve stability and effectiveness in multi-turn distillation.

\subsection{Trajectory-Level KL Instability in Multi-Turn On-Policy Distillation}


In this section, we conduct a pilot study on ALFWorld to examine the behavior of OPD in multi-turn settings. 
We systematically evaluate student–teacher pairs across the Qwen3 and Qwen2.5 model families, including both larger-scale and domain-adapted teachers. 
For Qwen3, we use Qwen3-30B-A3B-Instruct as the teacher and Qwen3-\{0.6, 1.7, 4\}B as students. 
For Qwen2.5, we adopt a GRPO-trained Qwen2.5-7B model as the teacher and Qwen2.5-\{0.5, 1.5, 3, 7\}B as students.

\begin{figure}[t]
\centering
\begin{subfigure}[b]{0.23\textwidth}
\centering
\includegraphics[width=\textwidth]{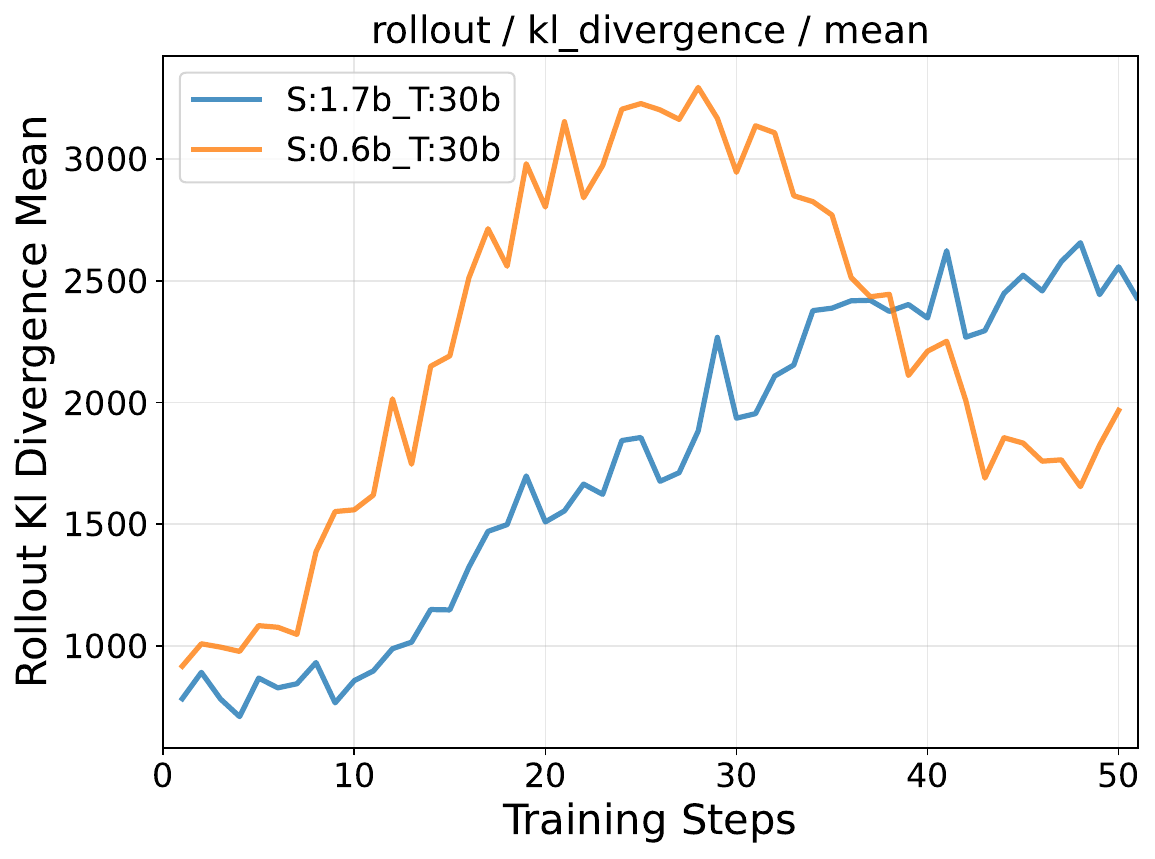}
\caption{Trajectory-level KL escalates during training.}
\label{fig:vanilla_kl}
\end{subfigure}
\hfill
\begin{subfigure}[b]{0.23\textwidth}
\centering
\includegraphics[width=\textwidth]{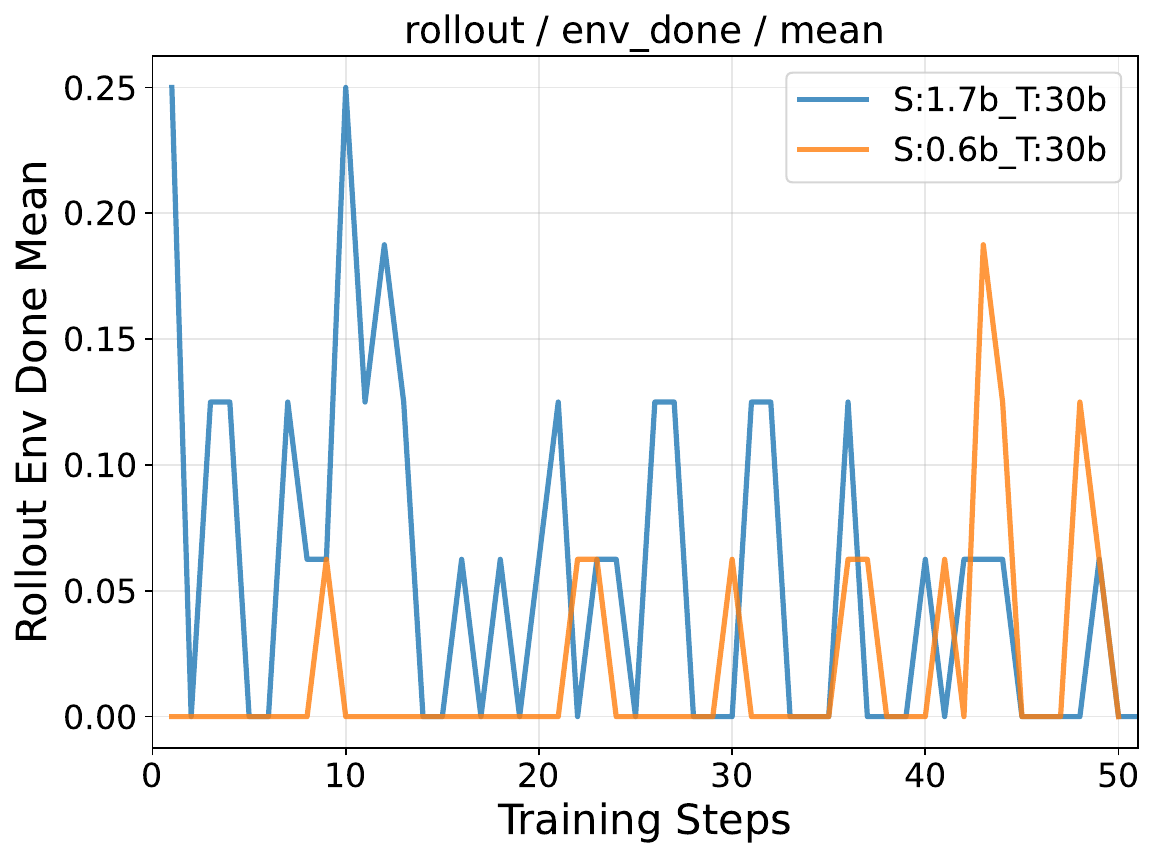}
\caption{Success rate collapses to zero as KL spikes.}
\label{fig:vanilla_success}
\end{subfigure}
\hfill
\begin{subfigure}[b]{0.23\textwidth}
\centering
\includegraphics[width=\textwidth]{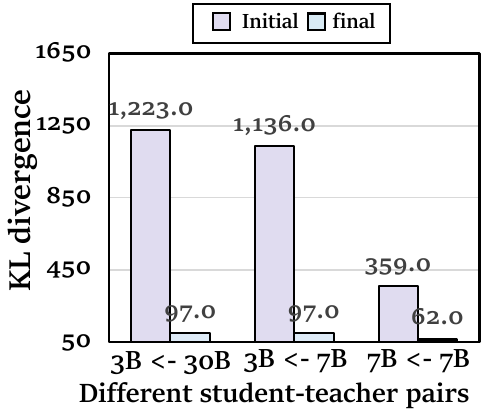}
\caption{Initial and final $KL$ during OPD training.}
\label{fig:kl bar}
\end{subfigure}
\hfill
\begin{subfigure}[b]{0.23\textwidth}
\centering
\includegraphics[width=\textwidth]{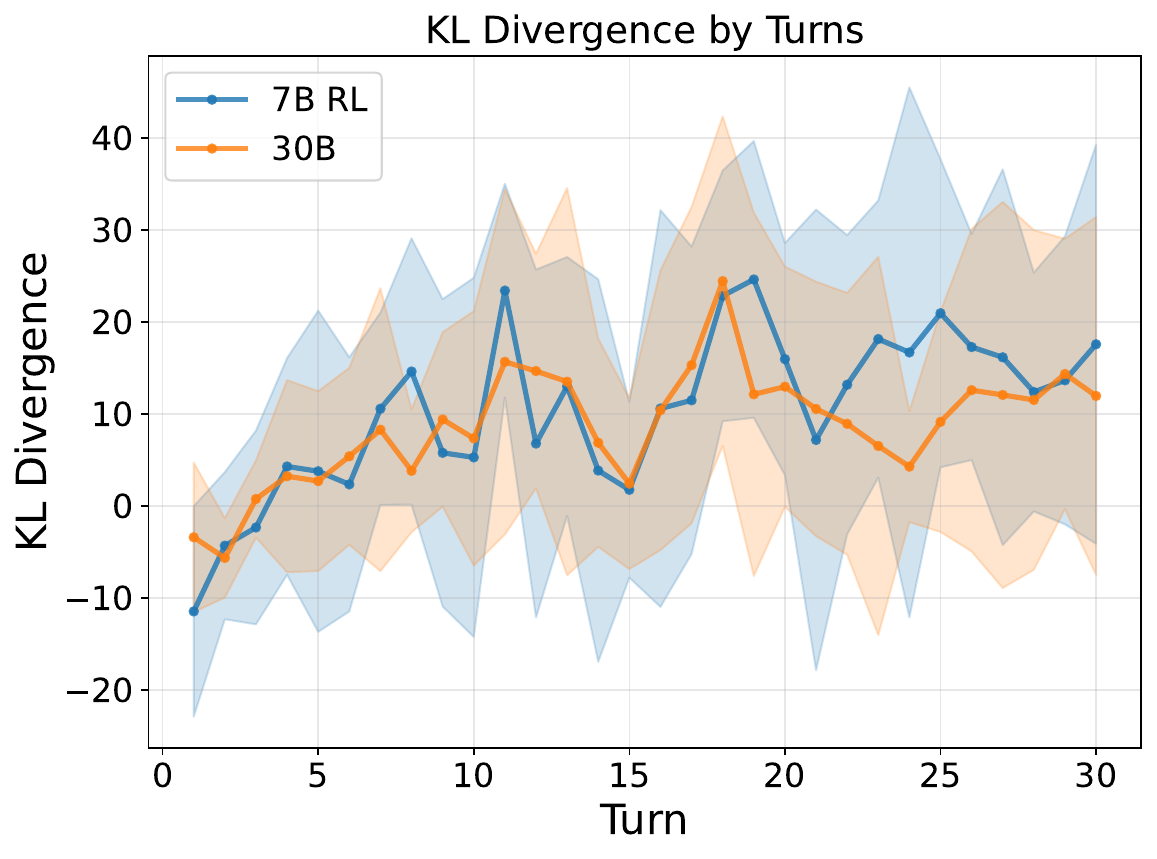}
\caption{Per-turn KL divergence increases as errors accumulate.}
\label{fig:per_turn_kl}
\end{subfigure}
\vspace{-0.5em}
\caption{
    \textbf{Trajectory-level KL analysis across different teacher--student pairs on ALFWorld.}
    (a)(b) show that the KL divergence escalates throughout training and task completion rates collapse. 
    (c) shows the large gap between the initial and converged KL divergence during OPD training. 
    (d) reveals the underlying reason: the KL divergence grows with the turn index, indicating compounding error amplification over the trajectory.
}
\vspace{-1.5em}
\label{fig:challenges}
\end{figure}


\textbf{Observation 1: KL escalation and success rate collapse co-occur during training.}
Unlike prior work on single-turn settings such as mathematics or question answering, 
where the KL divergence consistently converges and decreases throughout training, we observe that the KL divergence escalates with the number of training steps in multi-turn agent scenarios. 
As shown in Figure~\ref{fig:vanilla_kl}\&\ref{fig:vanilla_success}, when the student model (Qwen3-\{0.6,1.7\}B) is trained under vanilla OPD 
with a strong teacher (Qwen3-30B-A3B-Instruct), the trajectory-level KL divergence escalates rapidly, and the task success rate collapses to near-zero. 

\textbf{Observation 2: Although KL divergence converges, it suffers from a prohibitively high initial value.}
Moreover, we conduct experiments on different student models and observe that although their KL divergence eventually converges, they start with a prohibitively high value.
As shown in Figure~\ref{fig:kl bar}, across different student-teacher pairs (Qwen3-3B distilled from Qwen3-30B-A3B-Instruct, and Qwen2.5-\{3,7\}B distilled from a GRPO-trained Qwen2.5-7B model), we consistently observe the initial KL divergence ($\sim1000$) is typically orders of magnitude larger than its converged value ($\sim60$), indicating severe instability during the training for multi-turn OPD.
More details refer to Appendix~\ref{app: motivation}.

\textbf{The underlying mechanism: Compounding error amplification over the trajectory.}
The above two observations motivated us to investigate why directly applying OPD to agents leads to such KL escalation and training instability. 
To this end, in Figure~\ref{fig:per_turn_kl}, we visualize the per-turn KL divergence for Qwen2.5-3B distilled from a GRPO-trained Qwen2.5-7B and Qwen3-30B-A3B-Instruct, and observe a consistent increase with the turn index.

Regardless of whether the increasing KL divergence reflects the student’s inability to imitate the teacher or is a consequence of the student entering out-of-distribution states where the teacher becomes uncertain, the underlying issue remains the same—\textit{error accumulation over turns}. 
This is an inherent property of long-horizon multi-turn agents: student-generated actions and observations are appended to the history $h_t$, inducing causal coupling across turns and resulting in an increasing trend in KL divergence.
For small students, this is catastrophic; for larger ones, it is partially tolerated but remains highly inefficient.



\begin{remark}Note that Long-CoT increases the response length on the same environment state. 
However, multi-turn agents update the environment state at each interaction by incorporating new observations and actions, thereby amplifying compounding errors over the trajectory.
\end{remark}

The above observations and analysis pose a challenge: how can we retain the benefits of OPD’s dense signal while avoiding destabilization from accumulated errors in long-horizon interactions?
To address this, we turn to curriculum learning, where the model is first trained on easy problems and progressively exposed to hard ones. 

\subsection{Our Proposal: Temporal Curriculum On-Policy Distillation}
\label{sec:method}
\begin{figure}[t]
    \centering
    \includegraphics[width=\linewidth]{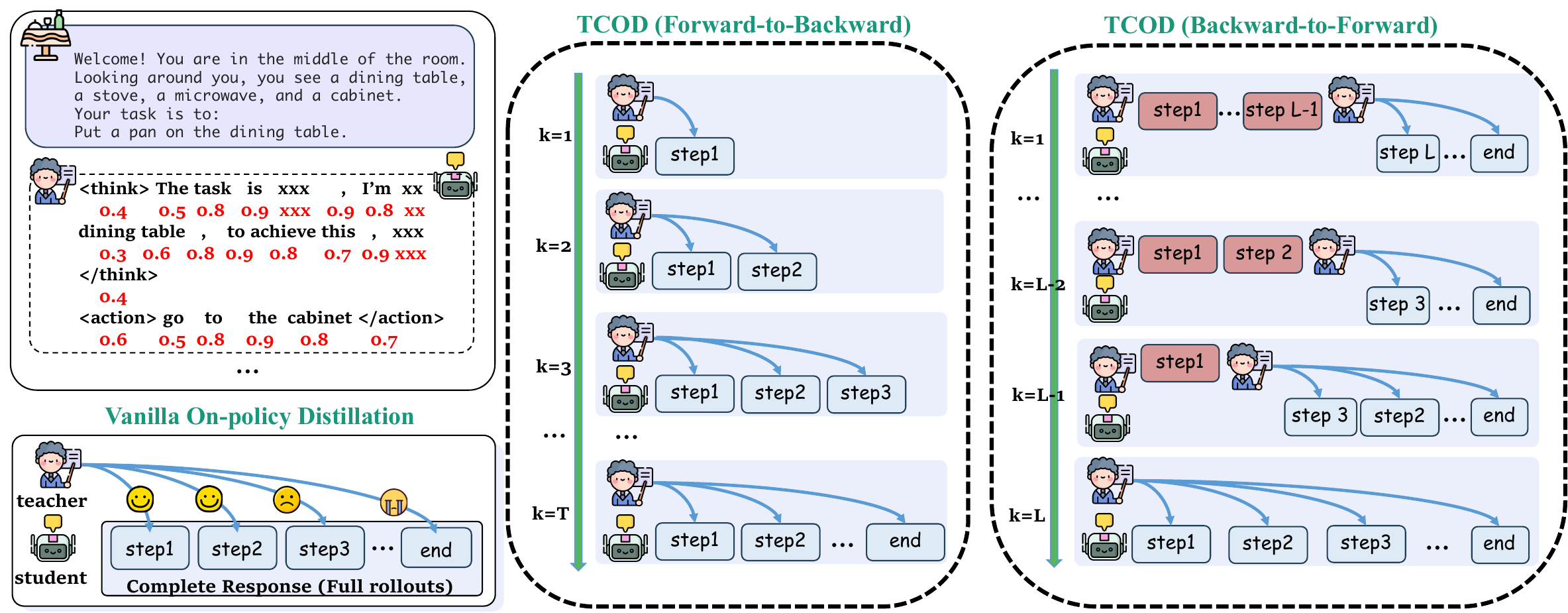}
    \caption{\textbf{Overview of our method \ours-F2B/B2F.} Comparison of vanilla on-policy distillation and \ours. \textbf{Left} is the OPD, \textbf{middle} is the illustration of \ours-F2B, and \textbf{right} is \ours-B2F. $k$ is the linear pacing control the trajectory length. The blue step is executed by the student, and the red step is executed by the teacher with a stop gradient.}
    \vspace{-1.5em}
    \label{fig:method}
\end{figure}



Building on the observations and insights from the previous section, we propose \textbf{T}emporal \textbf{C}urriculum \textbf{O}n-Policy \textbf{D}istillation~(\ours), a principled approach that controls the trajectory depth of agent interactions during the training process.
Specifically, we introduce two variants: \ours-F2B and \ours-B2F, which explicitly impose step constraints in forward and reverse curricula, respectively.

\textbf{Forward-to-Backward Induced \oursfull (\ours-F2B).}
We implement a "shallow-to-deep" curriculum by restricting the maximum interaction steps of a trajectory during the training process. 
As shown in Figure~\ref{fig:method}(middle), in our \ours-F2B, the student policy $\pi_\theta$ rolls out for maximum $k$ steps to finish the task, where k starts from a small number and progressively increases to a larger one, the objective is as follows:
\begin{equation}
\mathcal{L}_{TCOD\_F2B}(\theta) = \mathbb{E}_{\tau \sim \pi_\theta} \left[ \sum_{t=0}^{k-1} \mathcal{D}_{KL} \left( \pi_\phi(a_t|h_t) \parallel \pi_\theta(a_t|h_t) \right) \right],
\end{equation}
where the student first focuses on early-turn learning signals and then progressively completes the task end-to-end, mitigating compounding errors and preventing horizon-induced KL collapse.
However, determining the optimal step size and starting point is challenging, as different environments and models exhibit varying reasoning capabilities. 
To address this, we begin with linear pacing across the training step:
\begin{equation}
k = k_{\text{start}} + \lfloor n / \eta \rfloor, \quad n \in {1, \dots, N},
\label{eq:pacing}
\end{equation}
where $n$ represents the current training step and $N$ is the total number of training steps, $k_{\text{start}}$ defines the initial number of interaction steps and $\eta$ controls the curriculum’s growth rate. 
This approach requires only minor code changes. The whole algorithm is as follows:

\begin{algorithm}[h]
\caption{Temporal Curriculum On-Policy Distillation: \ours-F2B}
\label{alg:tcod_f2b}
\begin{algorithmic}[1]
\STATE \textbf{Input:} Student $\pi_\theta$, Teacher $\pi_\phi$, Environment $\mathcal{E}$,
       total steps $N$, curriculum parameters $k_{\text{start}}$, $\eta$
\STATE \textbf{Output:} Trained student policy $\pi_\theta$
\FOR{$n = 1, 2, \dots, N$}
    \STATE $k \leftarrow \min\!\left(k_{\text{start}} + \lfloor n / \eta \rfloor,\ T_{\max}\right)$
    \STATE Initialize $s_0 \sim \mathcal{E}$, history $h_0 \leftarrow \emptyset$
    \FOR{$t = 0, 1, \dots, k-1$}
        \STATE Sample $a_t \sim \pi_\theta(\cdot \mid h_t)$; execute $a_t$; update $h_{t+1}$
    \ENDFOR
    \STATE $\mathcal{L} \leftarrow \sum_{t=0}^{k} \mathcal{D}_{\mathrm{KL}}\!\left(\pi_\phi(a_t \mid h_t) \,\|\, \pi_\theta(a_t \mid h_t)\right)$
    \STATE Update $\theta \leftarrow \theta - \nabla_\theta \mathcal{L}$
\ENDFOR
\RETURN $\pi_\theta$
\end{algorithmic}
\end{algorithm}

Furthermore, to better exploit the teacher model, we propose \ours-B2F, which leverages the teacher to avoid early-turn error accumulation.

\textbf{Backward-to-Forward Induced \oursfull (\ours-B2F).}
In this variant, the teacher policy $\pi_\phi$ acts as a ``navigator.'' 
We initialize the environment to an intermediate state obtained by executing the initial prefix of a pre-collected successful trajectory $\tau^*$ using the teacher policy $\pi_\phi$, and let the agent start interaction from this state. 
Specifically, as shown in Figure~\ref{fig:method}, the teacher executes the first $L-k$ steps of its successful trajectory $\tau^*$ in the environment, after which the student policy $\pi_\theta$ takes over from this immediate state to continue planning and execution. The objective is as follows:
\begin{equation}
\mathcal{L}_{\text{TCOD\_B2F}}(\theta) = \mathbb{E}_{\tau \sim (\pi_\phi, \pi_\theta)} \left[ \sum_{t=L-k+1}^{T-1} \mathcal{D}_{KL} \left( \pi_\phi(a_t|h_t) \parallel \pi_\theta(a_t|h_t) \right) \right],
\end{equation}
where $L$ denotes the length of the successful trajectory $\tau^*$ for a given task, and $k$ is defined as in Equation~\ref{eq:pacing}, monotonically expanding until 
the student completes the task end-to-end throughout training.
This implementation is similarly lightweight, requiring only a simple warmup loop as shown in follows.
\begin{algorithm}[h]
\caption{Temporal Curriculum On-Policy Distillation: \ours-B2F}
\label{alg:tcod_b2f}
\begin{algorithmic}[1]
\STATE \textbf{Input:} Student $\pi_\theta$, Teacher $\pi_\phi$, Environment $\mathcal{E}$,
       total steps $N$, curriculum parameters $k_{\text{start}}$, $\eta$
\STATE \textbf{Output:} Trained student policy $\pi_\theta$
\STATE Pre-collect teacher successful trajectories $\mathcal{T}^* \leftarrow \{\tau^*\}$
\FOR{$n = 1, 2, \dots, N$}
    \STATE $k \leftarrow \min\!\left(k_{\text{start}} + \lfloor n / \eta \rfloor,\ L\right)$
    \STATE Sample $\tau^* \in \mathcal{T}^*$ with length $L$; initialize $s_0 \sim \mathcal{E}$
    \FOR{$t = 0, 1, \dots, L - k - 1$} 
        \STATE Execute teacher action $a_t^*$ \textbf{(stop gradient)}; update $h_{t+1}$
    \ENDFOR
    \FOR{$t = L-k, \dots, L$}
        \STATE Sample $a_t \sim \pi_\theta(\cdot \mid h_t)$; execute $a_t$; update $h_{t+1}$
    \ENDFOR
    \STATE $\mathcal{L} \leftarrow \sum_{t=L-k}^{L} \mathcal{D}_{\mathrm{KL}}\!\left(\pi_\phi(a_t \mid h_t) \,\|\, \pi_\theta(a_t \mid h_t)\right)$
    \STATE Update $\theta \leftarrow \theta - \nabla_\theta \mathcal{L}$
\ENDFOR
\RETURN $\pi_\theta$
\end{algorithmic}
\end{algorithm}

This mechanism effectively \textbf{bypasses compounding action errors} by ensuring the student only optimizes on trajectories initiated from successful, teacher-vetted prefixes. Crucially, the teacher steps of the trajectory do not contribute to the gradient, serving only to place the student on the ``doorstep of success.'' 
Detailed algorithms are provided in Appendix~\ref{appendix:algorithm}.

\textbf{Discussion of the train-test mismatch in \ours-B2F.}
During training, the student starts from a teacher-navigated checkpoint, whereas at test time it must act end-to-end from scratch. 
To this end, we gradually reduce the teacher's prefix from $L-1$ steps down to zero, ensuring that by the end of training the student executes the full trajectory from the initial state with no teacher intervention, fully aligning the training and 
test distributions. 
As shown in Appendix~\ref{app: b2f}, the end-to-end success rate on the test set increases steadily with training steps, confirming that the smooth curriculum transition effectively prevents catastrophic distribution shift in practice.

\subsection{Asynchronous Training Details for Stability}
\label{sec:implementation}

While the core \ours framework is conceptually straightforward, several practical design choices significantly impact training stability and efficiency in real-world deployments. 
All experiments were conducted on 8$\times$ NVIDIA H20 (96GB) GPUs. 
We describe our key implementation strategies:

\paragraph{Asynchronous Rollout and Training.}
To maximize GPU utilization, we decouple trajectory collection and 
model optimization into separate asynchronous processes. 
We use a pool of \textbf{actor} processes for rollout to continuously sample trajectories, while a 
central \textbf{learner} process for training uses these trajectories from a shared buffer and performs gradient updates. 
We use a lock-free ring buffer to minimize synchronization overhead. 
In our experiments, we allocate 4×H20 GPUs for actors and 2×H20 GPUs for learners.
Moreover, we use the remaining 2×H20 GPUs for the teachers.

\paragraph{Staleness-Aware Sub-trajectory Experience Replay.}
To maximize sample efficiency in multi-turn environments, we decompose each complete trajectory into a set of recursive sub-trajectories. Specifically, for a trajectory of length $n$, we store each prefix sequence $\tau_{1:t} = (s_0, a_0, \dots, s_t)$ as an independent experience entry in the replay buffer for $t \in \{1, \dots, n\}$. 
To prevent the input context from exceeding the model's effective memory limit, leading to training instability, 
we encapsulate the interaction history within the prompt as a structured context.
Consequently, the number of rollouts generated per batch is dynamic, depending on the varying lengths of collected trajectories. 
In our asynchronous setting, each trajectory is tagged with the version number $n$ of the policy $\pi_{\theta_n}$ used for collection. 
We implement a staleness filter that discards any experience where $n_{\text{current}} - n_{\text{old}} > \Delta_{\text{max}}$.
Empirically, we find that $\Delta_{\text{max}} = 2$ provides an optimal balance between sample efficiency and the strictness of the on-policy constraint.

\section{Experiments}

In this section, we conduct experiments on various benchmarks to evaluate our approach. Mainly, we design the experiments to study the following key questions:

\noindent{$\mathbf{Q1}$:} Compared to vanilla OPD, how does \ours 
alleviate KL escalation and recover the performance for small student models, and how does it enhance training stability and the performance for the larger ones?

\noindent{$\mathbf{Q2}$:} Can \ours enable the student to generalize effectively to tasks beyond the teacher's own capability boundary?

\noindent{$\mathbf{Q3}$:} How sensitive is \ours to the growth rate 
of the curriculum, and how does it compare to vanilla OPD in terms 
of training efficiency?

\subsection{Experimental Setup}

\begin{wraptable}{r}{0.5\textwidth}
\vspace{-10pt}
\centering
\scriptsize
\setlength{\tabcolsep}{1.5pt}
\caption{\textbf{Summary of the benchmarks used}.}
\vspace{-1em}
\begin{tabular}{lccccc}
\toprule
\textbf{Benchmark}  & \textbf{OOD} & \textbf{Type} & \textbf{Difficulty}  & \textbf{Max Turns} \\
\midrule
ALFWorld-seen & & Embodied & Easy & $30$ \\
ALFWorld-unseen & \cmark & Embodied & Medium & $30$ \\
ALFWorld-hard  & \cmark & Embodied & Hard  & $30$ \\
Webshop  & & E-commerce & Medium   & $15$ \\
ScienceWorld  & & Scientific Reason & Hard  & $30$ \\
\bottomrule
\end{tabular}
\label{tab:dataset_summary}
\vspace{-10pt}
\end{wraptable}


\textbf{Benchmarks.} 
We conduct experiments on three benchmarks including the embodied navigation environment ALFWorld~\citep{shridhar2020alfworld}, e-commerce platform WebShop~\citep{yao2022webshop}, and scientific reasoning ScienceWorld~\citep{wang2022scienceworld} as illustrated in Table~\ref{tab:dataset_summary}, spanning a spectrum of reasoning levels from simple to complex.
Max turns means the maximum exploration steps for each task.
For ALFWorld, we evaluate on both the \textit{seen} and 
\textit{unseen} splits, where the unseen split contains novel room 
layouts and object combinations not encountered during training, 
serving as our OOD evaluation.
We additionally construct a \textbf{Hard} set comprising 
tasks where the teacher fails under pass@10 sampling on the train split, to test 
whether \ours can generalize beyond the teacher's own capability 
boundary.
More benchmark details refer to Appendix~\ref{app: bench}.

\begin{table*}[t]
\centering
\caption{\textbf{Out-of-domain and hard set performance comparison} between \ours and OPD on ALFWorld. 
SR is success rate (\%) and Rounds is averge action rounds per task.
The best result is \textbf{bolded} and the second-best is \underline{underlined}. {\color{mygreen}Green} and {\color{myred}red} subscripts indicate improvement and degradation over Vanilla OPD, respectively.
}
\label{tab:main_alfworld}
\resizebox{\linewidth}{!}{
\begin{tabular}{lcccccc}
\toprule
\multirow{2}{*}{\textbf{Method}}
    & \multicolumn{2}{c}{\textbf{Valid Seen}}
    & \multicolumn{2}{c}{\textbf{Valid Unseen}}
    & \multicolumn{2}{c}{\textbf{Hard}} \\
\cmidrule(lr){2-3} \cmidrule(lr){4-5} \cmidrule(lr){6-7}
    & \textbf{SR} $\uparrow$ & \textbf{Rounds} $\downarrow$
    & \textbf{SR} $\uparrow$ & \textbf{Rounds} $\downarrow$
    & \textbf{SR} $\uparrow$ & \textbf{Rounds} $\downarrow$ \\
\midrule
\rowcolor{teachercolor}
Qwen2.5-7B-RL (Teacher)
    & 85.71 & 10.61
    & 76.87 & 13.06
    & 6.61  & 27.31 \\
\midrule
\rowcolor{groupcolor}
\multicolumn{7}{c}{\textit{Qwen2.5-3B}~{\small\textcolor{gray}{(Student)}}} \\
\quad Zero-Shot
    & 7.86  & 28.73
    & 2.24  & 29.63
    & 0.83  & 29.88 \\
\quad SFT
    & 32.14 & 22.85
    & 25.37 & 24.16
    & 4.96  & 29.12 \\
   
\quad Vanilla OPD
    & 65.72             & 14.73
    & 60.45             & 16.21
    & \underline{10.74} & 28.64 \\
    \midrule
\rowcolor{ourscolor}
\quad \ours~(B2F)
    & \underline{77.86}$_{\color{mygreen}\uparrow 12.14}$
    & \underline{12.57}$_{\color{mygreen}\downarrow 2.16}$
    & \underline{70.90}$_{\color{mygreen}\uparrow 10.45}$
    & \underline{14.56}$_{\color{mygreen}\downarrow 1.65}$
    & \textbf{13.22}$_{\color{mygreen}\uparrow 2.48}$
    & \textbf{28.16}$_{\color{mygreen}\downarrow 0.48}$ \\
\rowcolor{ourscolor}
\quad \ours~(F2B)
    & \textbf{81.43}$_{\color{mygreen}\uparrow 15.71}$
    & \textbf{11.76}$_{\color{mygreen}\downarrow 2.97}$
    & \textbf{79.19}$_{\color{mygreen}\uparrow 18.74}$
    & \textbf{12.47}$_{\color{mygreen}\downarrow 3.74}$
    & 9.92$_{\color{myred}\downarrow 0.82}$
    & \underline{28.57}$_{\color{mygreen}\downarrow 0.07}$ \\
\midrule
\rowcolor{groupcolor}
\multicolumn{7}{c}{\textit{Qwen2.5-7B}~{\small\textcolor{gray}{(Student)}}} \\
\quad Zero-Shot
    & 9.29  & 28.34
    & 8.96  & 28.46
    & 1.65  & 29.77 \\
\quad SFT
    & 54.29 & 18.92
    & 48.73 & 20.11
    & 8.26  & 28.73 \\
\quad Vanilla OPD
    & 75.37             & \underline{13.18}
    & 72.14             & 13.37
    & 13.22             & 27.89 \\
 
\midrule
\rowcolor{ourscolor}
\quad \ours~(B2F)
    & \textbf{86.43}$_{\color{mygreen}\uparrow 11.06}$
    & \textbf{11.06}$_{\color{mygreen}\downarrow 2.12}$
    & \textbf{77.61}$_{\color{mygreen}\uparrow 5.47}$
    & \textbf{13.16}$_{\color{mygreen}\downarrow 0.21}$
    & \textbf{20.66}$_{\color{mygreen}\uparrow 7.44}$
    & \textbf{27.07}$_{\color{mygreen}\downarrow 0.82}$ \\
\rowcolor{ourscolor}
\quad \ours~(F2B)
    & \underline{82.14}$_{\color{mygreen}\uparrow 6.77}$
    & 13.22$_{\color{myred}\uparrow 0.04}$
    & \underline{76.12}$_{\color{mygreen}\uparrow 3.98}$
    & \underline{13.22}$_{\color{mygreen}\downarrow 0.15}$
    & \underline{18.18}$_{\color{mygreen}\uparrow 4.96}$
    & \underline{27.37}$_{\color{mygreen}\downarrow 0.52}$ \\
\bottomrule
\end{tabular}
}
\end{table*}

\begin{figure*}[t]
\centering
\begin{subfigure}[b]{0.24\textwidth}
\centering
\includegraphics[width=\textwidth]{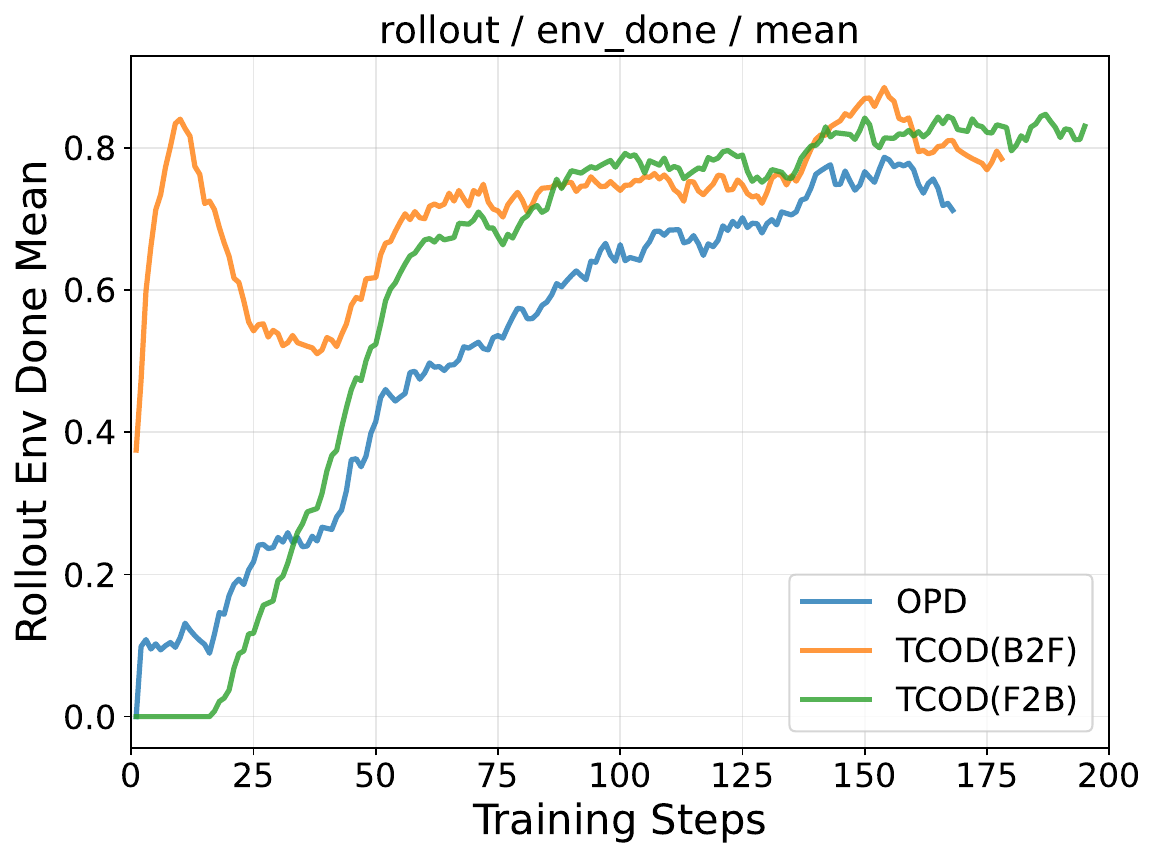}
\caption{Success Rates}
\label{fig:sr1}
\end{subfigure}
\hfill
\begin{subfigure}[b]{0.24\textwidth}
\centering
\includegraphics[width=\textwidth]{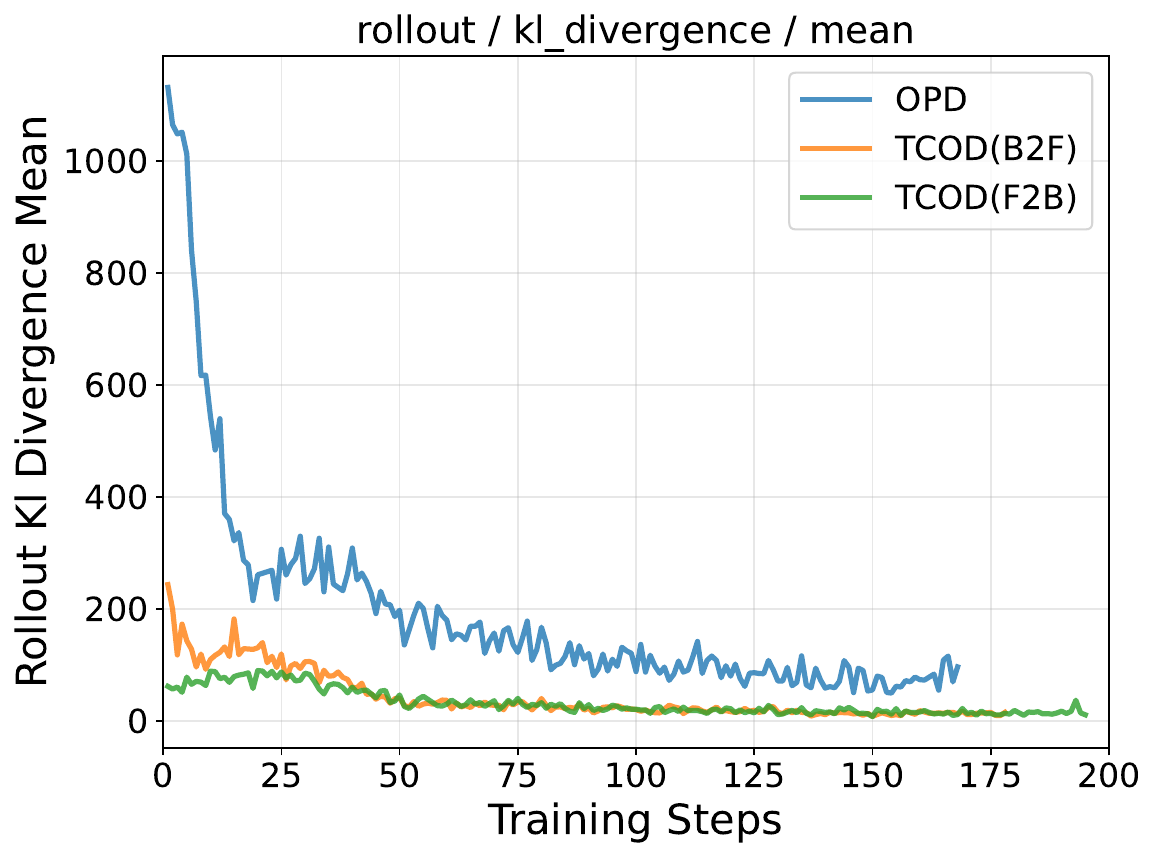}
\caption{KL Divergence}
\label{fig:kl1}
\end{subfigure}
\hfill
\begin{subfigure}[b]{0.24\textwidth}
\centering
\includegraphics[width=\textwidth]{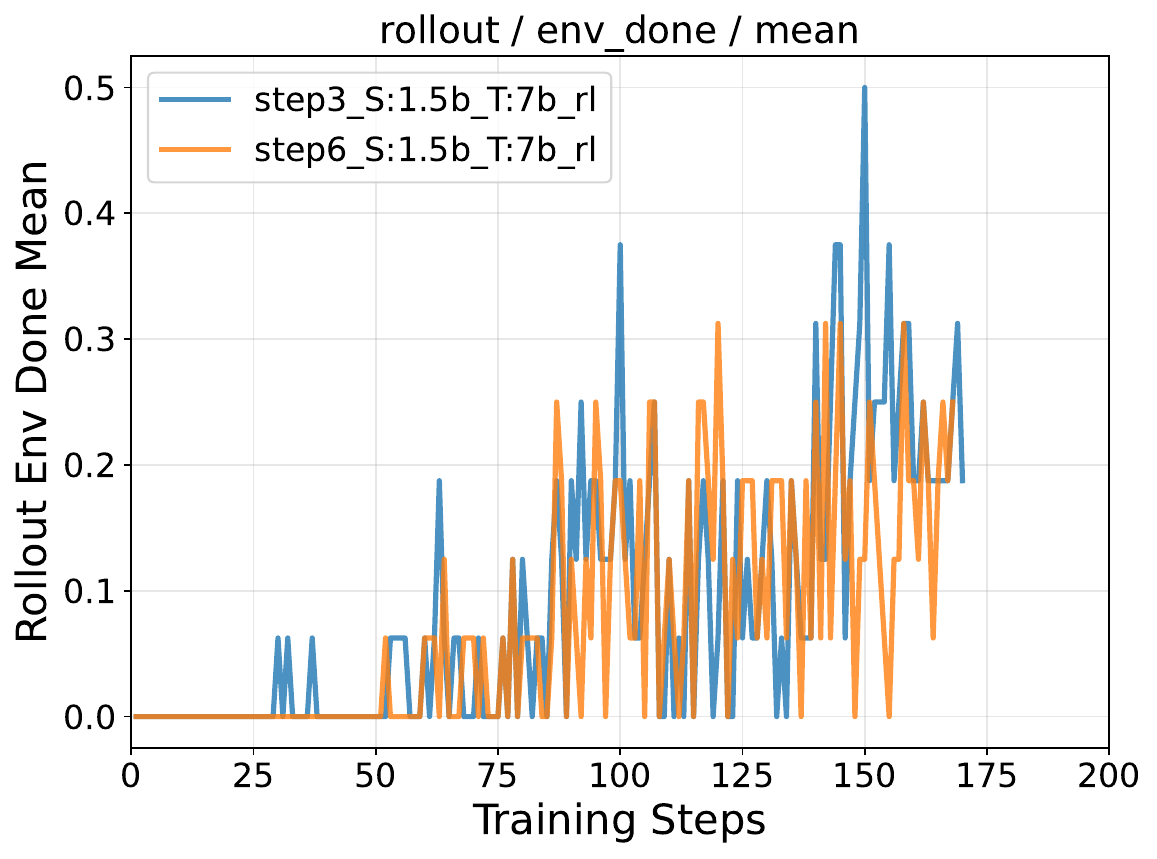}
\caption{Success Rates}
\label{fig:sr2}
\end{subfigure}
\hfill
\begin{subfigure}[b]{0.24\textwidth}
\centering
\includegraphics[width=\textwidth]{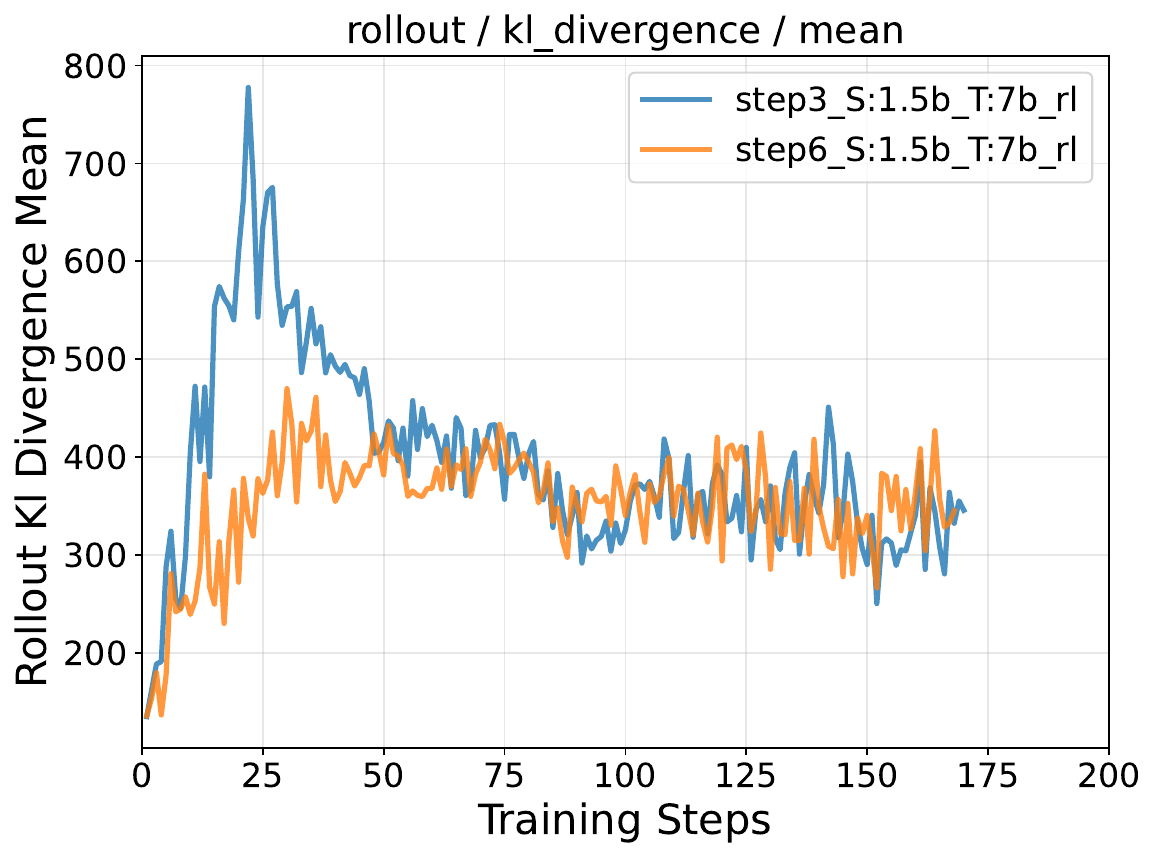}
\caption{KL Divergence}
\label{fig: kl2}
\end{subfigure}
\caption{
\textbf{Training Dynamics comparison of \ours and OPD on ALFWorld.} 
(a) and (b) show the success rate and KL divergence, respectively, for Qwen2.5-7B as the student. 
\ours maintains a higher success rate and more stable KL divergence. 
(c) and (d) show the success rate and KL divergence, respectively, for Qwen2.5-1.5B as the student model.
\ours-F2B under $\eta=3,6$ mitigate the success rate collapse and kl escalation.
}
\vspace{-1em}
\label{fig:training_dynamics_main}
\end{figure*}

\textbf{Training Details.}
For the main experiments on ALFWorld, we use Qwen2.5-3B and Qwen2.5-7B as student models, with Qwen2.5-7B fine-tuned via 
GRPO on the ALFWorld domain serving as the teacher.
For the cross-benchmark evaluation, we adopt Qwen3-1.7B and Qwen3-4B as students, with Qwen3-30B-A3B-Instruct as the teacher.
All experiments are conducted on 8$\times$ NVIDIA H20 GPUs.
We implement \ours based on the Reinforcement Fine-Tuning framework Trinity-RFT~\citep{pan2025trinityrftgeneralpurposeunifiedframework}. 
For expert trajectory collection for \ours-B2F initialization, we adopt a pass@10 sampling strategy using the teacher model, retaining only successful trajectories. 
For simplicity, we fix $k_{\text{start}} = 1$ and $\eta=2$ and examine the impact of different $\eta$ from $\{2, 4, 6\}$ in Sec~\ref{sec: ablation}. 


For baselines, we report the zero-shot student as the empirical lower bound and the teacher policy as the theoretical upper bound (Oracle). Moreover, we compare \ours with standard knowledge transfer paradigms, including supervised fine-tuning (SFT) and vanilla on-policy distillation (OPD). 
For evaluation, we test all the benchmarks using \textbf{success rate (SR)}, which measures the percentage of tasks completed successfully, where task completion is treated as a binary outcome. 
More details are provided in Appendix~\ref{appendix:hyperparameters}.



\subsection{$\mathbf{Q1}$: Alleviating KL Escalation and Improving Performance}
\label{subsec: performance}

\begin{table}[t]
\centering
\caption{%
    \textbf{Performance comparison between \ours-F2B/B2F and OPD.} 
We report Success Rate (\%) on validation sets. 
$\eta$ is the curriculum’s growth rate.
The best result is \textbf{bolded} and the second-best is \underline{underlined}. 
{\color{mygreen}Green} and {\color{myred}red} subscripts indicate improvement and degradation over Vanilla OPD, respectively.
}
\label{tab:cross_benchmark}
\resizebox{0.85\linewidth}{!}{
\begin{tabular}{lcccc}
\toprule
\textbf{Method}
    & \textbf{WebShop} $\uparrow$
    & \textbf{ALFWorld} $\uparrow$
    & \textbf{ScienceWorld} $\uparrow$
    & \textbf{Avg} $\uparrow$ \\
\midrule
\rowcolor{teachercolor}
Qwen3-30B (Teacher)
    & 32.84 & 39.57 & 18.42 & 30.28 \\
\midrule
\rowcolor{groupcolor}
\multicolumn{5}{c}{\textit{Qwen3-1.7B}~{\small\textcolor{gray}{(Student)}}} \\
\quad Vanilla OPD
    & 0.14  & 0.32  & 0.05  & 0.17  \\
\midrule
\rowcolor{b2fcolor}
\quad \ours~(B2F) $\eta$=2
    & \underline{20.54}$_{\color{mygreen}\uparrow 20.40}$
    & \underline{24.55}$_{\color{mygreen}\uparrow 24.23}$
    & \underline{10.82}$_{\color{mygreen}\uparrow 10.77}$
    & \underline{18.64}$_{\color{mygreen}\uparrow 18.47}$ \\
\rowcolor{b2fcolor}
\quad \ours~(B2F) $\eta$=4
    & \textbf{21.12}$_{\color{mygreen}\uparrow 20.98}$
    & 23.87$_{\color{mygreen}\uparrow 23.55}$
    & \textbf{11.34}$_{\color{mygreen}\uparrow 11.29}$
    & \textbf{18.78}$_{\color{mygreen}\uparrow 18.61}$ \\
\rowcolor{b2fcolor}
\quad \ours~(B2F) $\eta$=6
    & 20.33$_{\color{mygreen}\uparrow 20.19}$
    & \textbf{24.91}$_{\color{mygreen}\uparrow 24.59}$
    & 10.65$_{\color{mygreen}\uparrow 10.60}$
    & 18.63$_{\color{mygreen}\uparrow 18.46}$ \\
\midrule
\rowcolor{f2bcolor}
\quad \ours~(F2B) $\eta$=2
    & \underline{21.15}$_{\color{mygreen}\uparrow 21.01}$
    & \underline{24.12}$_{\color{mygreen}\uparrow 23.80}$
    & \underline{10.45}$_{\color{mygreen}\uparrow 10.40}$
    & \underline{18.57}$_{\color{mygreen}\uparrow 18.40}$ \\
\rowcolor{f2bcolor}
\quad \ours~(F2B) $\eta$=4
    & 20.44$_{\color{mygreen}\uparrow 20.30}$
    & \textbf{25.03}$_{\color{mygreen}\uparrow 24.71}$
    & 9.22$_{\color{mygreen}\uparrow 9.17}$
    & 18.23$_{\color{mygreen}\uparrow 18.06}$ \\
\rowcolor{f2bcolor}
\quad \ours~(F2B) $\eta$=6
    & \textbf{21.78}$_{\color{mygreen}\uparrow 21.64}$
    & 23.65$_{\color{mygreen}\uparrow 23.33}$
    & \textbf{11.08}$_{\color{mygreen}\uparrow 11.03}$
    & \textbf{18.84}$_{\color{mygreen}\uparrow 18.67}$ \\
\midrule
\rowcolor{groupcolor}
\multicolumn{5}{c}{\textit{Qwen3-4B}~{\small\textcolor{gray}{(Student)}}} \\
\quad Vanilla OPD
    & 30.12 & 36.85 & 15.95 & 27.64 \\
\midrule
\rowcolor{b2fcolor}
\quad \ours~(B2F) $\eta$=2
    & \textbf{32.15}$_{\color{mygreen}\uparrow 2.03}$
    & \underline{38.62}$_{\color{mygreen}\uparrow 1.77}$
    & \underline{17.46}$_{\color{mygreen}\uparrow 1.51}$
    & \textbf{29.41}$_{\color{mygreen}\uparrow 1.77}$ \\
\rowcolor{b2fcolor}
\quad \ours~(B2F) $\eta$=4
    & 29.21$_{\color{myred}\downarrow 0.91}$
    & 37.84$_{\color{mygreen}\uparrow 0.99}$
    & \textbf{16.73}$_{\color{mygreen}\uparrow 0.78}$
    & 27.93$_{\color{mygreen}\uparrow 0.29}$ \\
\rowcolor{b2fcolor}
\quad \ours~(B2F) $\eta$=6
    & 29.05$_{\color{myred}\downarrow 1.07}$
    & \textbf{39.35}$_{\color{mygreen}\uparrow 2.50}$
    & 15.88$_{\color{myred}\downarrow 0.07}$
    & \underline{28.09}$_{\color{mygreen}\uparrow 0.45}$ \\
\midrule
\rowcolor{f2bcolor}
\quad \ours~(F2B) $\eta$=2
    & \textbf{31.81}$_{\color{mygreen}\uparrow 1.69}$
    & \textbf{38.95}$_{\color{mygreen}\uparrow 2.10}$
    & \textbf{17.85}$_{\color{mygreen}\uparrow 1.90}$
    & \textbf{29.54}$_{\color{mygreen}\uparrow 1.90}$ \\
\rowcolor{f2bcolor}
\quad \ours~(F2B) $\eta$=4
    & \underline{30.54}$_{\color{mygreen}\uparrow 0.42}$
    & 37.62$_{\color{mygreen}\uparrow 0.77}$
    & \underline{17.23}$_{\color{mygreen}\uparrow 1.28}$
    & \underline{28.46}$_{\color{mygreen}\uparrow 0.82}$ \\
\rowcolor{f2bcolor}
\quad \ours~(F2B) $\eta$=6
    & 29.87$_{\color{myred}\downarrow 0.25}$
    & \underline{37.88}$_{\color{mygreen}\uparrow 1.03}$
    & 17.12$_{\color{mygreen}\uparrow 1.17}$
    & 28.29$_{\color{mygreen}\uparrow 0.65}$ \\
\bottomrule
\end{tabular}
}
\end{table}

\begin{figure}[t]
\centering
\begin{subfigure}[b]{0.24\textwidth}
\centering
\includegraphics[width=\textwidth]{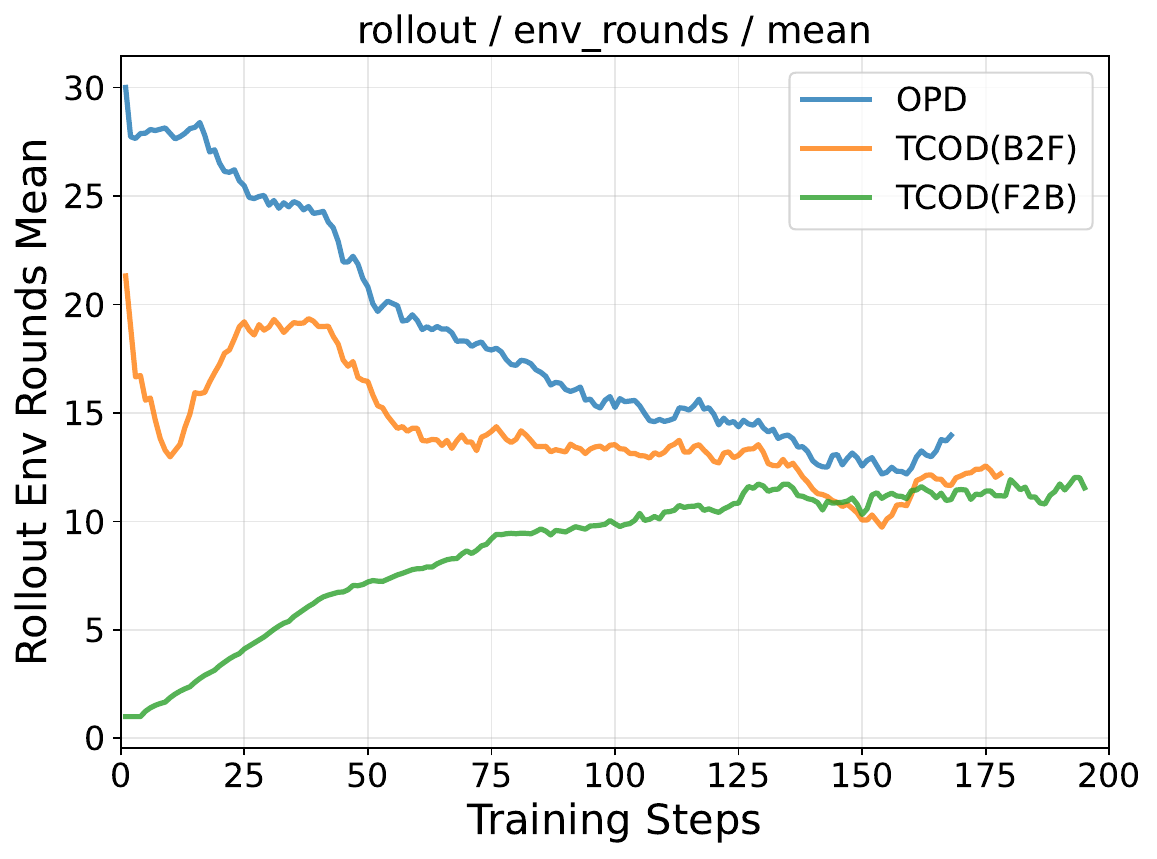}
\caption{Action Rounds}
\label{fig:rounds_training}
\end{subfigure}
\hfill
\begin{subfigure}[b]{0.24\textwidth}
\centering
\includegraphics[width=\textwidth]{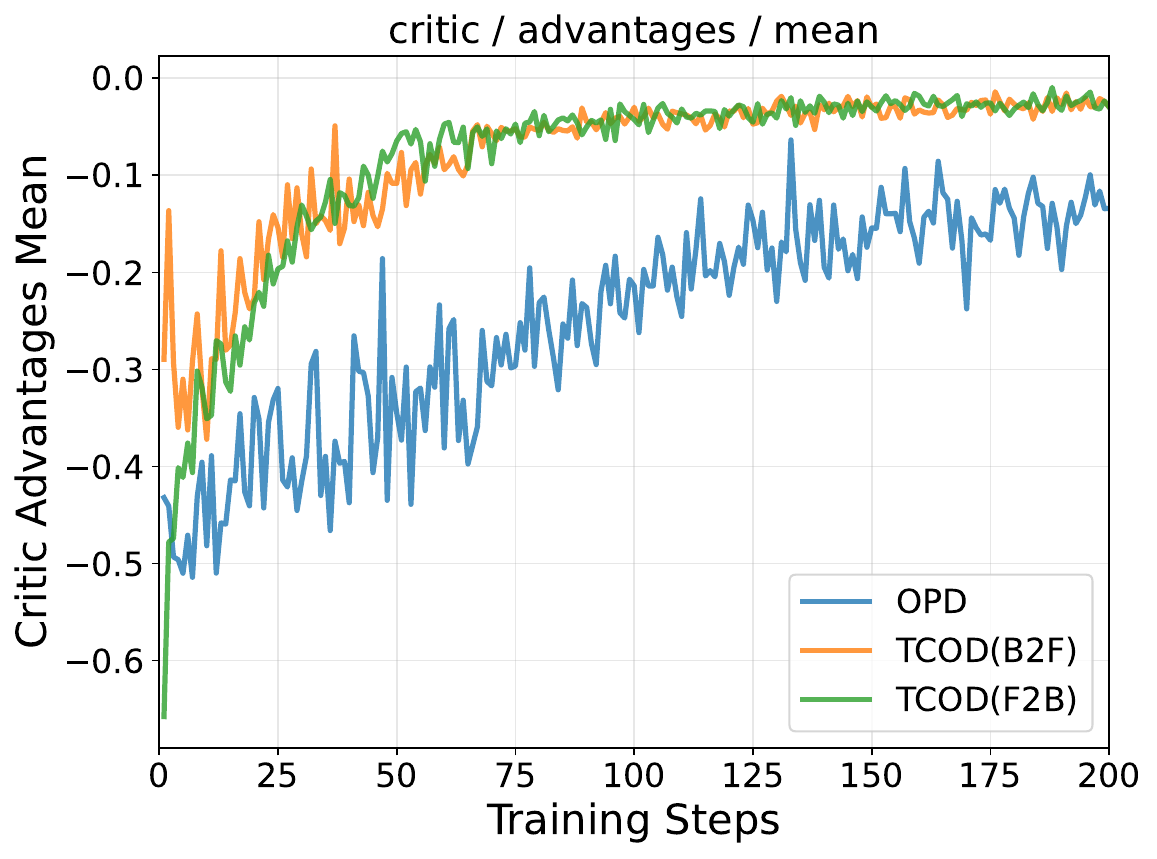}
\caption{Advantages}
\label{fig:advantage}
\end{subfigure}
\hfill
\begin{subfigure}[b]{0.24\textwidth}
\centering
\includegraphics[width=\textwidth]{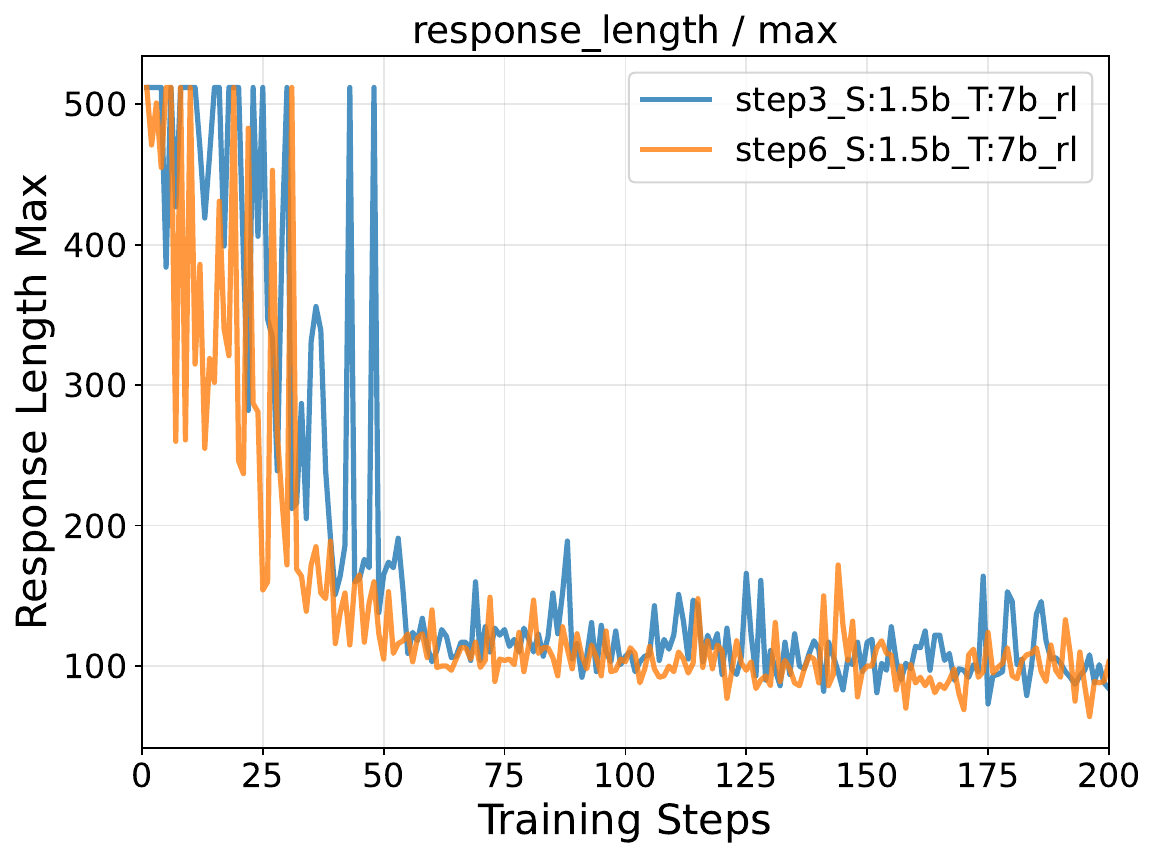}
\caption{Max Response length}
\label{fig:max length}
\end{subfigure}
\hfill
\begin{subfigure}[b]{0.24\textwidth}
\centering
\includegraphics[width=\textwidth]{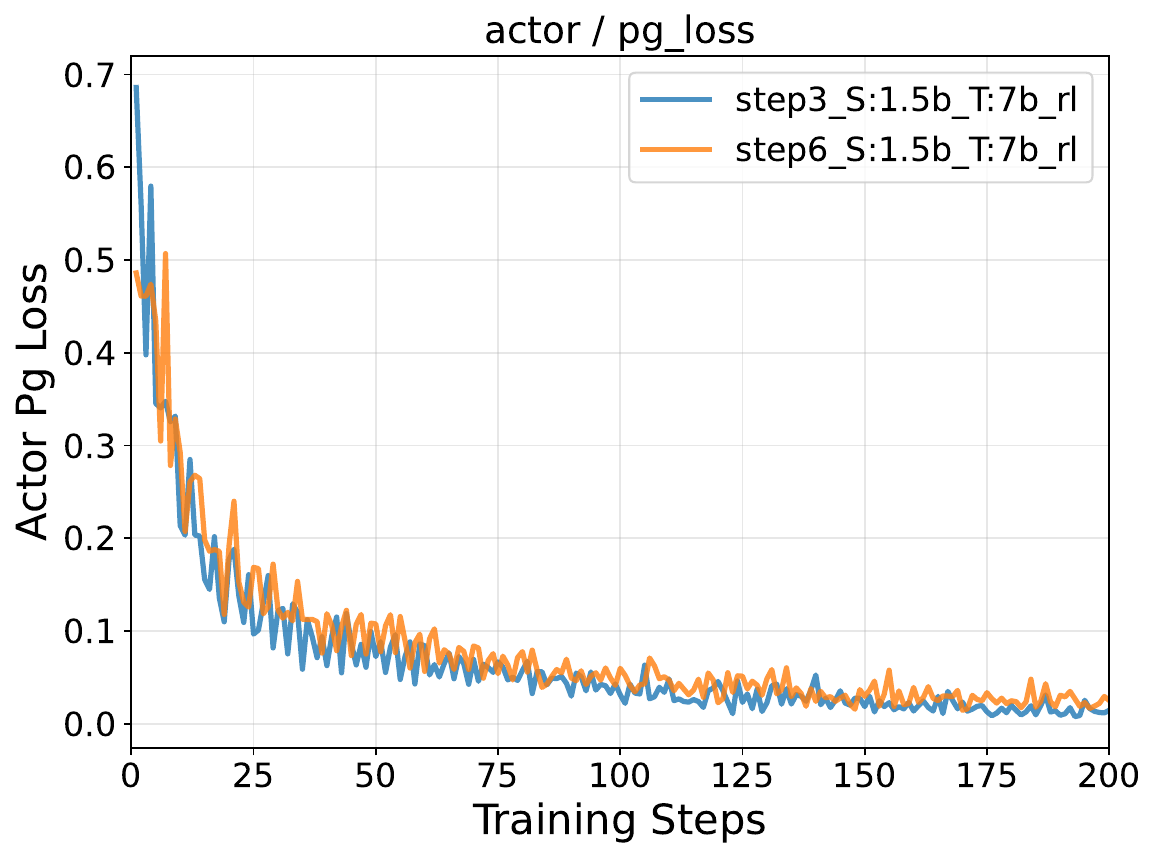}
\caption{Policy Gradient Loss}
\label{fig:gradient_norm}
\end{subfigure}
\caption{\textbf{Further Analysis of \ours-F2B/B2F on ALFWorld.} 
(a)(b) is the average action rounds,  advantages during training for Qwen2.5-7B as the student.
\ours effectively reduces the action rounds and achieves faster advantage convergence.
(c)(d) is the maximum response length, policy gradient loss during training for Qwen2.5-1.5B as the student model.
\ours mitigates redundant responses while maintaining training stability. 
}
\vspace{-1em}
\label{fig:challenges}
\end{figure}

In Table~\ref{tab:main_alfworld}, we present results of \ours on ALFWorld with students (Qwen2.5-3B, Qwen2.5-7B) and a GRPO-trained Qwen2.5-7B teacher, reporting both success rate (SR) and average action steps.
We find that \ours-F2B and B2F substantially outperform vanilla OPD and SFT across model scales. 
Notably, \ours reduces the average number of action steps by $2.97$ steps, while improving SR by up to 15.71 over OPD, suggesting that \textbf{curriculum learning from the teacher over trajectories leads to better performance}.
Figure~\ref{fig:sr1}\&\ref{fig:kl1}\&\ref{fig:advantage} further show that \ours achieves faster convergence in success rate and advantage, while maintaining more stable KL divergence than vanilla OPD.

\textbf{Different Benchmarks and Model Sizes.}
In Table~\ref{tab:cross_benchmark}, we evaluate \ours across three benchmarks using students Qwen3-1.7B and Qwen3-4B with a Qwen3-30B-A3B-Instruct teacher.
Overall, \ours-F2B and \ours-B2F achieve comparable performance to vanilla OPD.
Moreover, 
as illustrated in Figure~\ref{fig:sr2}\&\ref{fig: kl2}, \ours-F2B under both $\eta=\{3,6\}$ \textbf{maintains stable KL throughout training and achieves an increasing success rate, effectively mitigating KL escalation and improving average success rate by 18.67.}
Furthermore, Figure~\ref{fig:max length}\&\ref{fig:gradient_norm} shows additional training metrics, where \ours can recover from an explosion in response length, while the policy gradient loss decreases smoothly.

\subsection{$\mathbf{Q2}$: Generalizing Beyond the Teacher's Capability Boundary}
\label{subsec: beyond}

Beyond the performance gains and KL stability achieved by \ours, 
we further investigate whether \ours can enable the student to surpass the teacher itself.
Table~\ref{tab:main_alfworld} reports performance on both the unseen environment split and the hard split.
Specifically, the hard split comprises 121 challenging tasks from ALFWorld where the teacher performs poorly.
On the unseen split, \ours already outperforms the teacher by up to 2.5 points in SR.
More surprisingly, on the Train Hard split, both \ours-B2F and \ours-F2B substantially exceed the teacher's SR of 6.61, with \ours-B2F achieving a gain of up to 14 points.
This demonstrates that \textbf{\ours does not merely imitate the teacher, but develops a more robust policy that generalizes beyond the teacher's capability boundary.}

\subsection{$\mathbf{Q3}$: Robustness, Sensitivity, and Efficiency Analysis of \ours}
\label{sec: ablation}

\textbf{Curriculum’s Growth Rate $\eta$ Ablation.}
Table~\ref{tab:cross_benchmark} reports the effect of varying the curriculum’s growth rate $\eta \in \{2, 4, 6\}$ across different benchmarks. 
Performance remains consistently stronger than vanilla OPD across settings, with less than 2\% variation in success rate, demonstrating that \ours-F2B/B2F is not sensitive to the specific choice of $\eta$. 
This robustness makes \ours easy to deploy in practice without extensive hyperparameter tuning.
Nonetheless, as shown in Figure~\ref{fig: kl2}, a larger $\eta$ leads to more stable KL divergence during training, as the student spends more iterations mastering the current trajectory depth before the curriculum advances to longer horizons. 
In practice, we recommend starting with a small $\eta$ to allow the curriculum to progress quickly in the early stages, and increasing $\eta$ if the KL divergence instability during training is observed.

\textbf{Domain-Specific \textit{vs.} Larger Teacher.}
Comparing Table~\ref{tab:main_alfworld} and Table~\ref{tab:cross_benchmark}, we find that teacher quality strongly affects the upper bound of \ours.
In Table~\ref{tab:main_alfworld}, the teacher is a GRPO-tuned Qwen2.5-7B on ALFWorld, reaching 85.71\% success rate. Under this setting, \ours-B2F with the same 7B backbone even slightly surpasses the teacher by 0.7 points.
In Table~\ref{tab:cross_benchmark}, the teacher is Qwen3-30B-A3B-Instruct, a general model with weaker performance on the target domain. In this case, both vanilla OPD and \ours fail to exceed the teacher, with about a 2-point gap.
This suggests that the teacher’s performance on the target domain matters more than model scale alone in enabling student improvement.

\begin{wrapfigure}{r}{0.45\textwidth} 
\centering
\includegraphics[width=0.45\textwidth]{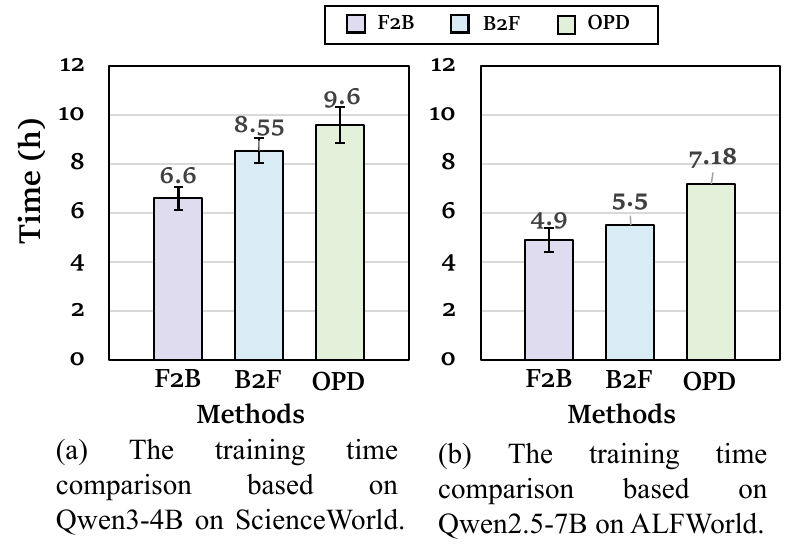} 
\vspace{-2em}
\caption{\textbf{Training time comparison}.}
\label{fig:time}

\end{wrapfigure}

\textbf{\ours is computationally efficient.}
Figure~\ref{fig:time} compares the total training cost of \ours and vanilla OPD on ALFWorld and ScienceWorld.
On both benchmarks, \ours-F2B and \ours-B2F reduce total training time by nearly 32\% compared to vanilla OPD. 
This gain comes from the step-based curriculum in \ours: early in training, the student takes fewer steps, producing shorter trajectories and faster data collection.
Notably, \ours-F2B is more efficient than \ours-B2F.
This is because \ours-F2B limits the maximum interaction steps to $k$, while \ours-B2F, though starting from intermediate states, still leads the student to take extra exploratory actions, producing longer trajectories.
Figure~\ref{fig:rounds_training} further verifies that \ours-F2B uses fewer rollout action steps than \ours-B2F, and both require fewer steps than vanilla OPD.

\section{Conclusion}

In this work, we identify a fundamental failure mode of vanilla OPD in multi-turn agents, termed Trajectory-Level KL Instability, where compounding errors across turns lead to escalating KL divergence and unreliable teacher supervision. 
Building on this insight, we propose \ours, a simple and principled framework that controls the trajectory depth exposed to the student during training, instantiated through two practical variants: Forward-to-Backward (F2B) and Backward-to-Forward (B2F).
Extensive experiments demonstrate that \ours consistently stabilizes training, recovers small models from collapse, improves success rates for larger models, and reduces total training time compared to vanilla OPD.
Beyond practical improvements, \ours opens new directions for curriculum-guided training of long-horizon autonomous agents.



\bibliographystyle{colm2026_conference}
\bibliography{colm2026_conference}
\newpage
\appendix
\begin{center}
	\LARGE \bf {Appendix}
\end{center}

\etocdepthtag.toc{mtappendix}
\etocsettagdepth{mtchapter}{none}
\etocsettagdepth{mtappendix}{subsubsection}
\tableofcontents
\newpage

\section{Limitations and Future Work}
While \ours offers practical benefits, it also comes with a few limitations that point to interesting future work. 
\ours-B2F relies on pre‑collected successful teacher trajectories, which may require additional trajectory collection overhead. In such cases, the forward‑to‑backward variant (\ours‑F2B) provides a drop‑in alternative that requires no demonstrations.
Although we empirically observe that \ours's fixed curriculum schedule is robust across the three benchmarks and model sizes we studied, the optimal pace may vary with different environments or student–teacher pairs. 
An adaptive mechanism that automatically adjusts the horizon based on the student’s learning progress—such as through an exponential moving average of the KL divergence—could further improve generality; we consider this a promising direction for future investigation.
Our evaluation focuses on three text‑based multi‑turn benchmarks; extending \ours to multimodal or physically embodied environments is an important next step to assess its generality. 
These considerations do not compromise \ours's practical effectiveness, but instead highlight promising directions for further improvement.

\section{Additional Observation}
\label{app: motivation}

We systematically evaluate student–teacher pairs across the Qwen3 and Qwen2.5 model families, including both larger-scale and domain-adapted teachers. 
For Qwen3, we use Qwen3-30B-A3B-Instruct as the teacher and Qwen3-\{0.6, 1.7, 4\}B as students. 
For Qwen2.5, we adopt a GRPO-trained Qwen2.5-7B model as the teacher and Qwen2.5-\{0.5, 1.5, 3, 7\}B as students.

\begin{figure}[t]
\centering
\begin{subfigure}[b]{0.32\textwidth}
\centering
\includegraphics[width=\textwidth]{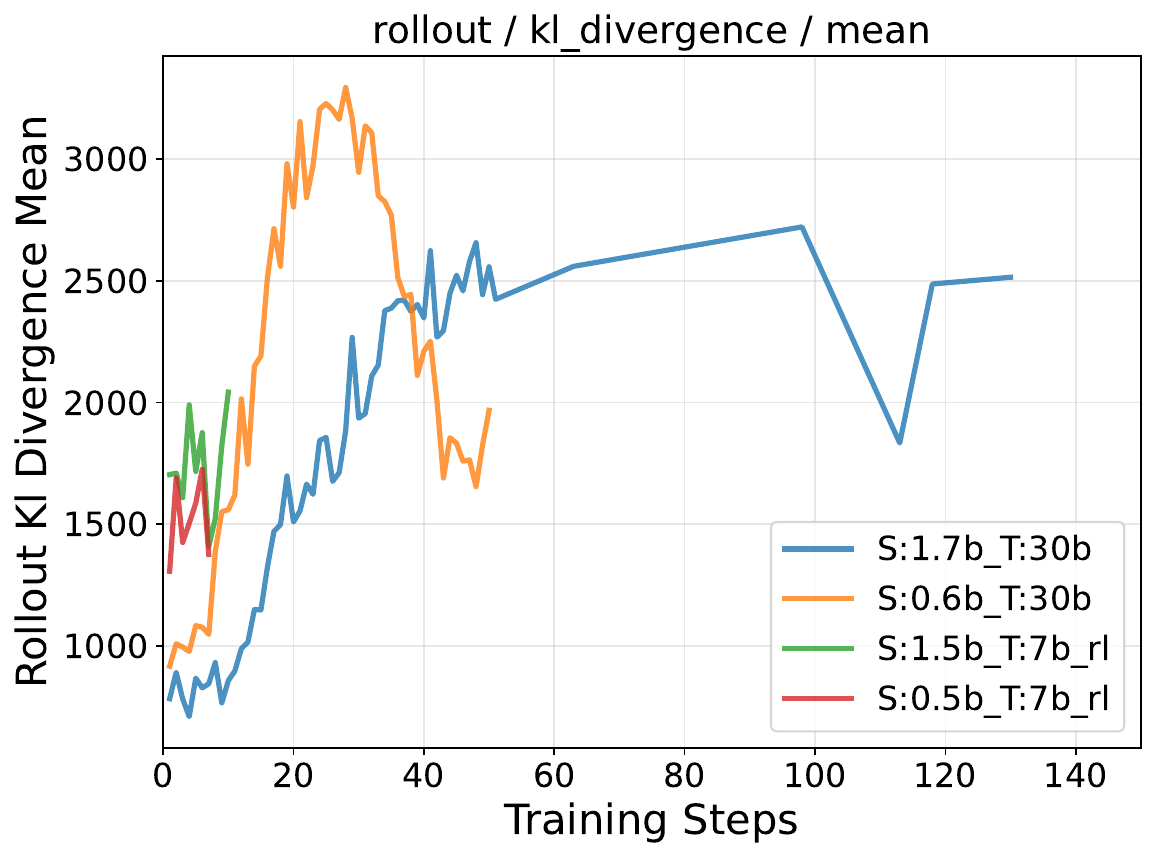}
\caption{Trajectory-level KL escalates during training.}
\label{app:vanilla_kl}
\end{subfigure}
\hfill
\begin{subfigure}[b]{0.32\textwidth}
\centering
\includegraphics[width=\textwidth]{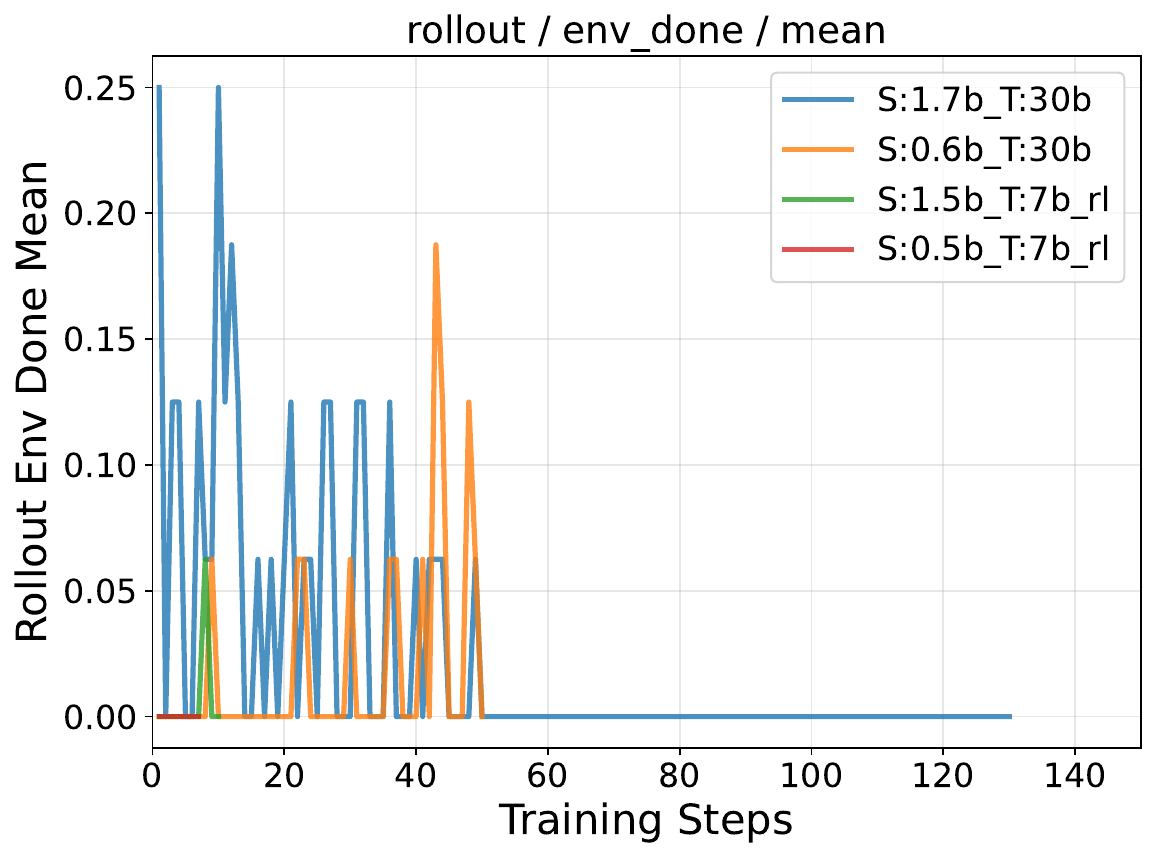}
\caption{Success rate collapses to zero as KL spikes.}
\label{app:vanilla_success}
\end{subfigure}
\hfill
\begin{subfigure}[b]{0.32\textwidth}
\centering
\includegraphics[width=\textwidth]{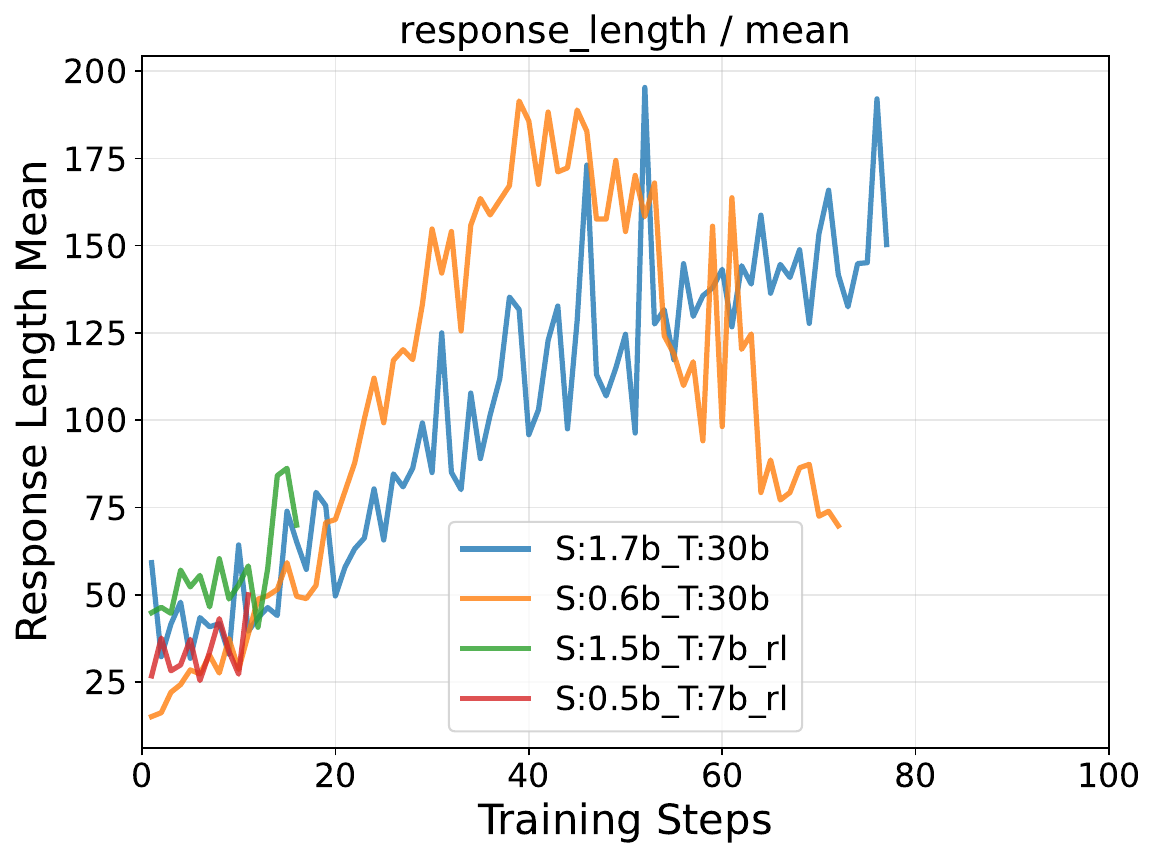}
\caption{Response length increases across turns.}
\label{app:per_turn_kl}
\end{subfigure}

\caption{
    \textbf{KL Escalation and success rate across Teacher--Student Pairs.}
    We evaluate Qwen3-\{0.6B, 1.7B\} (teacher: Qwen3-30B-A3B-Instruct) and Qwen2.5-\{0.5B, 1.5B\} 
    (teacher: Qwen2.5-7B-RL) under vanilla OPD on ALFWorld.
}
\label{app:challenges}
\end{figure}

\paragraph{Observation 1: KL escalation and success rate collapse co-occur in small models ($<$3B).}
Unlike prior work in single-turn settings (e.g., math or QA), where the KL divergence typically decreases and stabilizes during training, we observe a fundamentally different behavior in multi-turn agent environments. As shown in Figure~\ref{app:challenges}, when training small student models (Qwen3-{0.6B, 1.7B} and Qwen2.5-{0.5B, 1.5B}) with vanilla OPD, the trajectory-level KL divergence increases sharply as training progresses. This escalation is accompanied by a simultaneous collapse of the success rate to nearly zero. Moreover, response lengths grow steadily across turns, indicating compounding errors and increasingly off-distribution trajectories. Together, these results suggest that, in multi-turn settings, small models fail to maintain alignment with the teacher under their own rollout distribution, leading to unstable training dynamics and ineffective supervision.

\begin{figure}[t]
\centering
\begin{subfigure}[b]{0.32\textwidth}
\centering
\includegraphics[width=\textwidth]{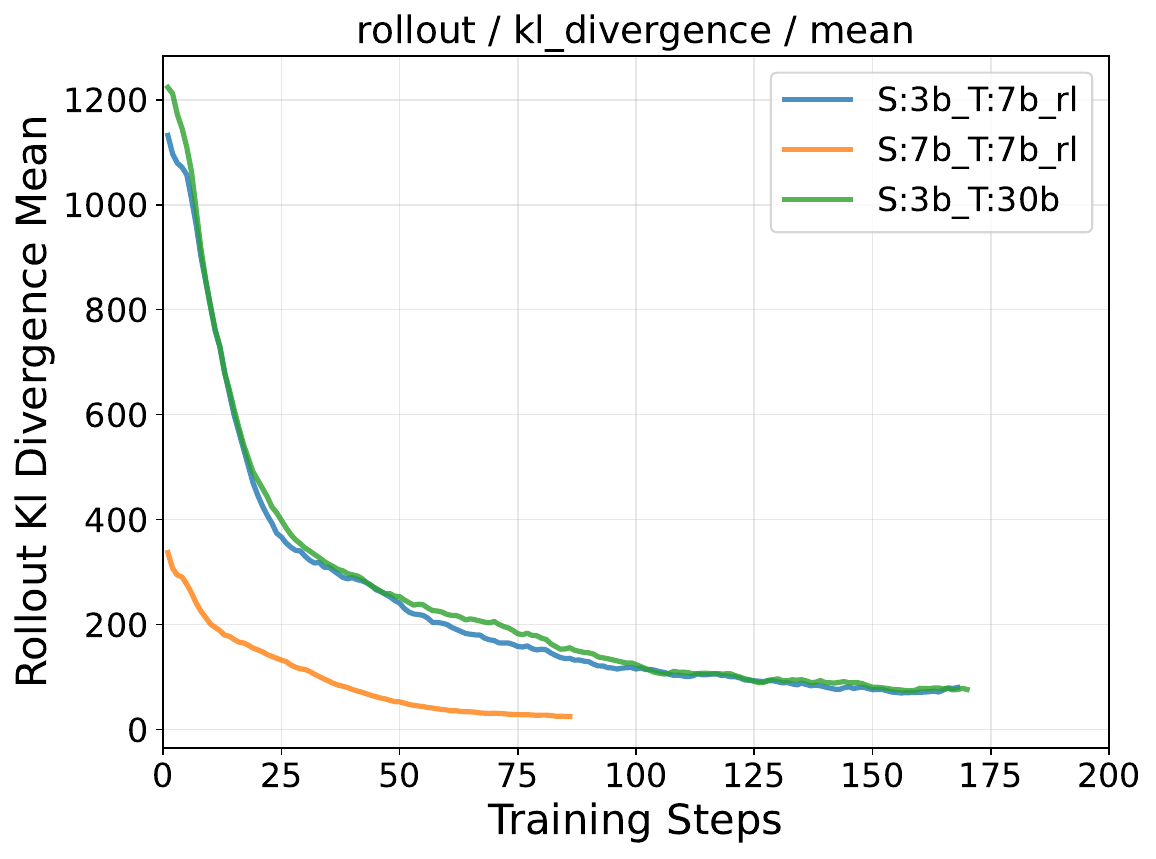}
\caption{KL divergence.}
\label{app:vanilla_kl}
\end{subfigure}
\hfill
\begin{subfigure}[b]{0.32\textwidth}
\centering
\includegraphics[width=\textwidth]{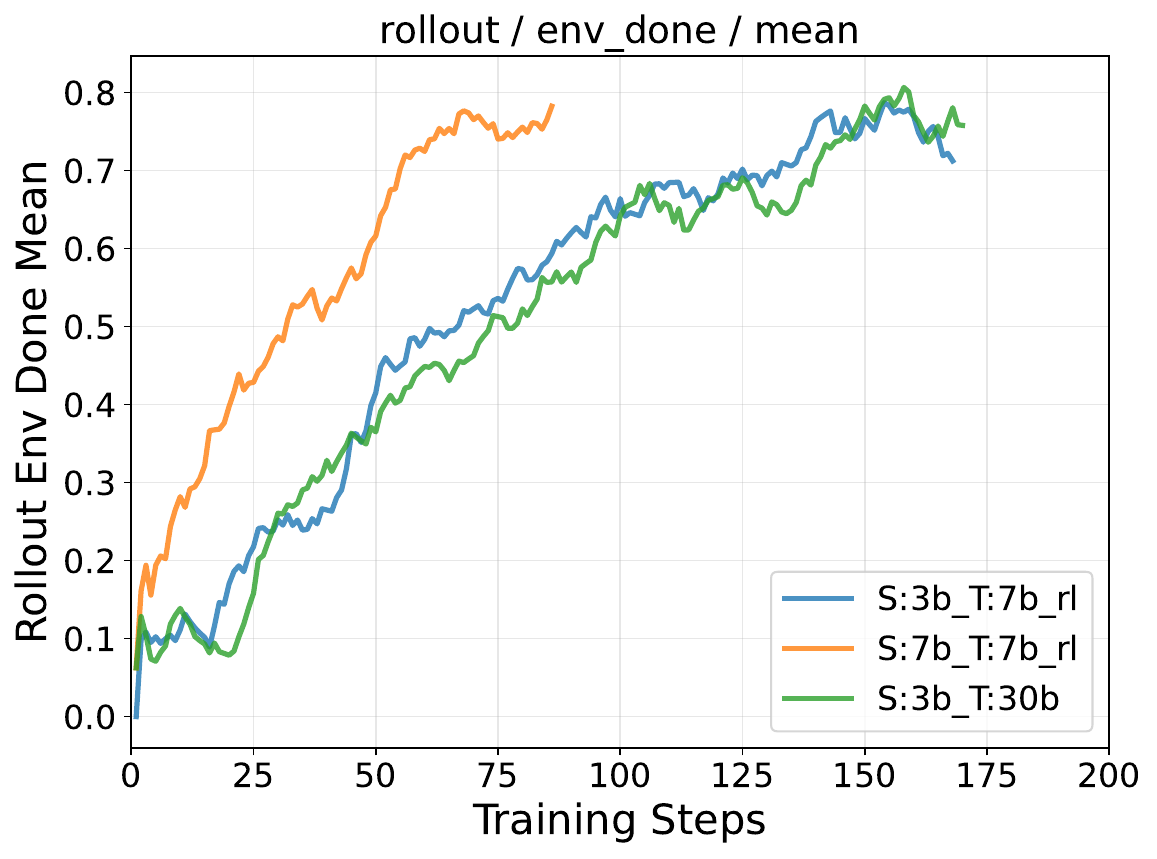}
\caption{Success rate.}
\label{app:vanilla_success}
\end{subfigure}
\hfill
\begin{subfigure}[b]{0.32\textwidth}
\centering
\includegraphics[width=\textwidth]{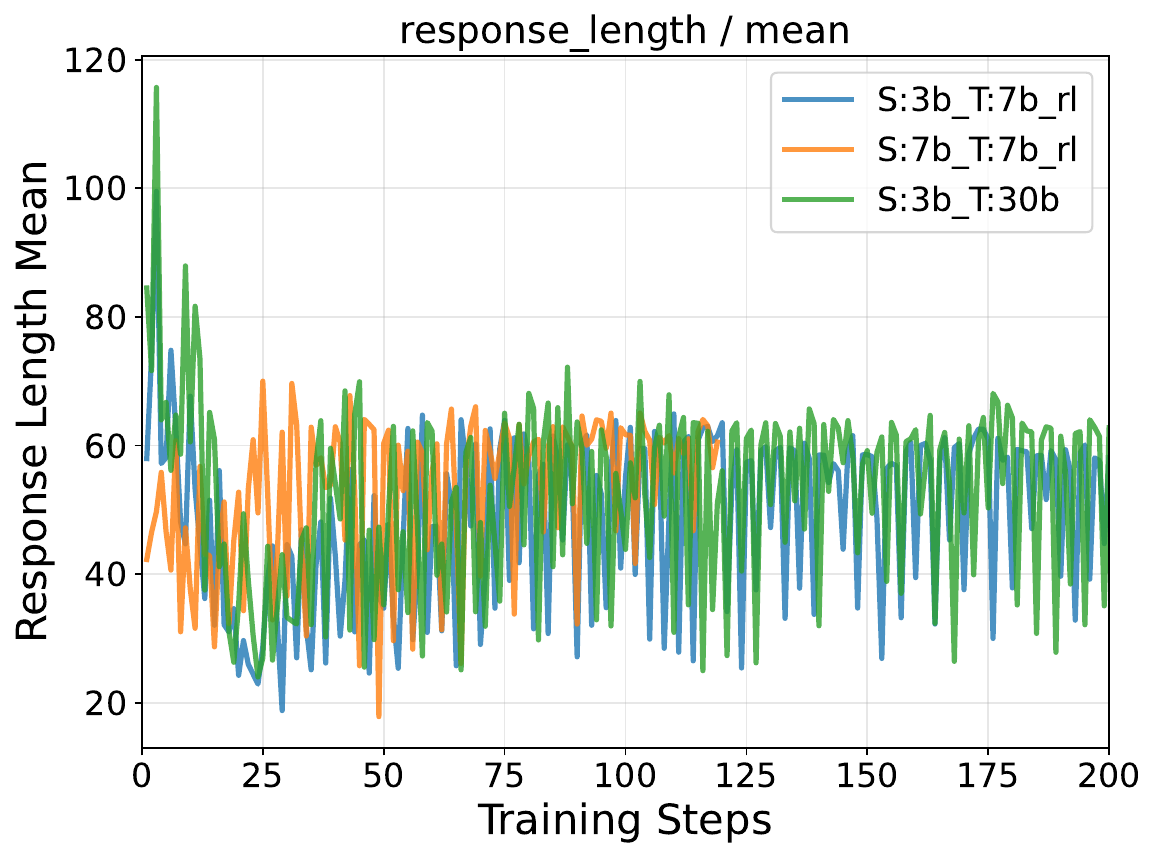}
\caption{Response length.}
\label{app:per_turn_kl}
\end{subfigure}

\caption{
    \textbf{Horizon-Induced KL Escalation across Teacher--Student Pairs.}
    We evaluate Qwen2.5-\{3B, 7B\} 
    (teacher: Qwen3-30B-A3B-Instruct, Qwen2.5-7B-RL) under vanilla OPD on ALFWorld.
}
\label{app:challenges2}
\end{figure}

\paragraph{Observation 2: Teacher–student matching matters; stronger teachers are not always better.}
We further examine the impact of teacher–student pairing in Figure~\ref{app:challenges2}. For a 3B student, training under both a strong 30B teacher and a 7B RL teacher leads to similar outcomes: the KL divergence decreases steadily and the success rate improves at comparable rates, indicating that increasing teacher strength beyond a certain point does not yield additional benefits. In contrast, when the student capacity matches the teacher more closely (7B student with 7B RL teacher), the KL divergence converges significantly faster and the success rate rises more rapidly, outperforming both 3B student settings. This suggests that an appropriate capacity match between teacher and student is more critical than absolute teacher strength; overly strong teachers do not necessarily improve, and may even limit, distillation efficiency in multi-turn settings.

\section{Algorithm for \ours-F2B/B2F}
\label{appendix:algorithm}

Algorithm~\ref{alg:tcod_f2b} and Algorithm~\ref{alg:tcod_b2f} present the complete training procedures for \ours-F2B and \ours-B2F, respectively, integrating the curriculum pacing strategy and implementation details described in Section~\ref{sec:method}.

In \ours-F2B (Algorithm~\ref{alg:tcod_f2b}), the student policy $\pi_\theta$ rolls out the trajectory for $k$ steps at each training iteration, where $k$ is progressively expanded according to the linear pacing schedule in Equation~\ref{eq:pacing}. By concentrating the distillation signal on early-turn states at the beginning of training and gradually extending the horizon, the student builds a robust foundation before being exposed to the full trajectory, effectively mitigating compounding errors and preventing KL collapse.

\begin{algorithm}[h]
\caption{Temporal Curriculum On-Policy Distillation: \ours-F2B}
\label{alg:tcod_f2b}
\begin{algorithmic}[1]
\STATE \textbf{Input:} Student $\pi_\theta$, Teacher $\pi_\phi$, Environment $\mathcal{E}$,
       total steps $N$, curriculum parameters $k_{\text{start}}$, $\eta$
\STATE \textbf{Output:} Trained student policy $\pi_\theta$
\FOR{$n = 1, 2, \dots, N$}
    \STATE $k \leftarrow \min\!\left(k_{\text{start}} + \lfloor n / \eta \rfloor,\ T_{\max}\right)$
    \STATE Initialize $s_0 \sim \mathcal{E}$, history $h_0 \leftarrow \emptyset$
    \FOR{$t = 0, 1, \dots, k-1$}
        \STATE Sample $a_t \sim \pi_\theta(\cdot \mid h_t)$; execute $a_t$; update $h_{t+1}$
    \ENDFOR
    \STATE $\mathcal{L} \leftarrow \sum_{t=0}^{k} \mathcal{D}_{\mathrm{KL}}\!\left(\pi_\phi(a_t \mid h_t) \,\|\, \pi_\theta(a_t \mid h_t)\right)$
    \STATE Update $\theta \leftarrow \theta - \nabla_\theta \mathcal{L}$
\ENDFOR
\RETURN $\pi_\theta$
\end{algorithmic}
\end{algorithm}

In \ours-B2F (Algorithm~\ref{alg:tcod_b2f}), the teacher policy $\pi_\phi$ first replays the initial $L - k$ steps from a pre-collected successful trajectory $\tau^*$ without contributing to the gradient, placing the student at a vetted checkpoint state. The student then takes over for the remaining $k$ steps, learning to complete the task from progressively earlier starting points as $k$ increases. By the end of training, the teacher prefix is fully eliminated ($k = L$), ensuring the student executes the complete trajectory end-to-end and fully closing the train-test distribution gap.

\begin{algorithm}[h]
\caption{Temporal Curriculum On-Policy Distillation: \ours-B2F}
\label{alg:tcod_b2f}
\begin{algorithmic}[1]
\STATE \textbf{Input:} Student $\pi_\theta$, Teacher $\pi_\phi$, Environment $\mathcal{E}$,
       total steps $N$, curriculum parameters $k_{\text{start}}$, $\eta$
\STATE \textbf{Output:} Trained student policy $\pi_\theta$
\STATE Pre-collect teacher successful trajectories $\mathcal{T}^* \leftarrow \{\tau^*\}$
\FOR{$n = 1, 2, \dots, N$}
    \STATE $k \leftarrow \min\!\left(k_{\text{start}} + \lfloor n / \eta \rfloor,\ L\right)$
    \STATE Sample $\tau^* \in \mathcal{T}^*$ with length $L$; initialize $s_0 \sim \mathcal{E}$
    \FOR{$t = 0, 1, \dots, L - k - 1$} 
        \STATE Execute teacher action $a_t^*$ \textbf{(stop gradient)}; update $h_{t+1}$
    \ENDFOR
    \FOR{$t = L-k, \dots, L$}
        \STATE Sample $a_t \sim \pi_\theta(\cdot \mid h_t)$; execute $a_t$; update $h_{t+1}$
    \ENDFOR
    \STATE $\mathcal{L} \leftarrow \sum_{t=L-k}^{L} \mathcal{D}_{\mathrm{KL}}\!\left(\pi_\phi(a_t \mid h_t) \,\|\, \pi_\theta(a_t \mid h_t)\right)$
    \STATE Update $\theta \leftarrow \theta - \nabla_\theta \mathcal{L}$
\ENDFOR
\RETURN $\pi_\theta$
\end{algorithmic}
\end{algorithm}

\section{Experiment Details}
\label{appendix:hyperparameters}

\subsection{Benchmark Environments}
\label{app: bench}
\textbf{ALFWorld}~\citep{shridhar2020alfworld} is a text-based embodied environment requiring navigation and object manipulation across six categories of household tasks. 
ALFWorld provides \textit{seen} and \textit{unseen} splits: the seen split tests performance in environments present during training, while the unseen split requires the agent to operate in novel room layouts and object combinations, serving as our OOD evaluation.
For ALFWorld, we further build a \textbf{Hard} set of 121 tasks where the teacher fails under pass@10 sampling on the training split. This set serves as a more challenging OOD evaluation to test whether \ours can generalize beyond the teacher's own capability boundary.

\textbf{Webshop}~\citep{yao2022webshop} is a web-based environment requiring the agent to search and select products that match a given user instruction across multi-turn interactions with a simulated e-commerce platform.

\textbf{ScienceWorld}~\citep{wang2022scienceworld} is a text-based environment that tests scientific reasoning across 30 task types aligned with the elementary science curriculum. The agent receives a score between 0 and 100 at the end of each task based on task completion.

\subsection{Baselines}

To rigorously assess the effectiveness of \ours, we benchmark against the following paradigms, establishing clear performance boundaries for the student models:

\paragraph{Teacher (Upper Bound):} The performance of the expert policy ($\pi_{\phi}$) is evaluated directly on the environment. In standard distillation, this represents the theoretical upper bound, as the primary goal is to recover this capability within the smaller student model. 
Notably, our evaluation on the Train Hard split (Sec~\ref{subsec: beyond}) investigates whether TCOD can even generalize beyond this upper limit.

\paragraph{Zero-Shot Student (Lower Bound):} The base student model ($\pi_{\theta}$) evaluated directly on the interactive tasks without any task-specific fine-tuning or distillation. This establishes the absolute starting point of the student's reasoning capability in the agentic environments.

\textbf{Supervised Fine-Tuning (SFT):} The fundamental imitation learning baseline. The student model is fine-tuned via standard negative log-likelihood (NLL) loss strictly on the successful trajectories ($\tau^*$) pre-collected from the Teacher for 2 epochs, suffering from the well-known exposure bias in multi-turn settings.

\textbf{Vanilla On-Policy Distillation (OPD):} The standard multi-turn adaptation of recent OPD methods. The student is trained to minimize the token-level KL divergence against the teacher's distribution over the student's entire generated trajectory (full rollouts), without any horizon constraints or temporal curriculum. This serves as the direct baseline to demonstrate the Trajectory-Level KL Instability.

\subsection{Training Hyperparameters}

We conduct training across three text-based interactive environments: ALFWorld, ScienceWorld, and WebShop. The training configuration is summarized in Table~\ref{tab:train_params}.

\begin{table}[h]
\centering
\caption{Training hyperparameters for TCOD across all environments.}
\label{tab:train_params}
\small
\begin{tabular}{ll}
\toprule
\textbf{Hyperparameter} & \textbf{Value} \\
\midrule
\textbf{Algorithm} & \\
\quad Algorithm type & On-Policy Distillation \\
\quad Advantage function & Multi-turn OPD \\
\quad KL coefficient & 1.0 \\
\quad Learning rate & $1 \times 10^{-6}$ \\
\quad Gradient clipping & 1.0 \\
\quad Repeat times & 1 \\
\quad Sample strategy & Staleness control (max staleness: 2) \\
\midrule
\textbf{Training} & \\
\quad Total training steps & 250 \\
\quad Batch size & 16 \\
\quad Train batch size & 64 \\
\quad Save interval & 250 \\
\quad Evaluation interval & 5 steps \\
\midrule
\textbf{Model Configuration} & \\
\quad Max prompt tokens & 10,240 \\
\quad Max response tokens & 512 \\
\midrule
\textbf{Inference (Rollout)} & \\
\quad Temperature (training) & 1.0 \\
\quad Temperature (evaluation) & 0.4 \\
\quad Logprobs & Enabled (all tokens) \\
\quad Seed & 42 \\
\midrule
\textbf{Environment-Specific} & \\
\quad ALFWorld max steps & 30 \\
\quad ScienceWorld max steps & 30 \\
\quad WebShop max steps & 15 \\
\midrule
\textbf{TCOD Curriculum} & \\
\quad Workflow name & B2F, F2B \\
\quad Starting step & 1 \\
\quad Checkpoint steps & 2, 4, 6 \\
\midrule
\textbf{Distributed Training} & \\
\quad Number of nodes & 1 \\
\quad GPUs per node & 8 \\
\quad Tensor parallel size & 2 \\
\quad Sequence parallel size & 2 (Ulysses) \\
\quad Max tokens per GPU & 16,384 \\
\quad GPU memory utilization & 0.7 \\
\quad Data type & BFloat16 \\
\bottomrule
\end{tabular}
\end{table}

\subsection{Evaluation Hyperparameters}

For evaluation, we assess model performance on three test sets: \texttt{test\_unseen}, \texttt{test}, and \texttt{train\_hard} (ALFWorld only). The evaluation hyperparameters are consistent across all environments as shown in Table~\ref{tab:eval_params}.

\begin{table}[h]
\centering
\caption{Evaluation hyperparameters for TCOD across all environments.}
\label{tab:eval_params}
\small
\begin{tabular}{ll}
\toprule
\textbf{Hyperparameter} & \textbf{Value} \\
\midrule
\textbf{Generation} & \\
\quad Maximum tokens & 4,096 \\
\quad Temperature & 0.4 (evaluation) \\
\quad Top-p & 1.0 \\
\quad Top-k & -1 \\
\quad MinP & 0.0 \\
\midrule
\textbf{Environment} & \\
\quad Max environment steps & 30 (ALFWorld, ScienceWorld) / 15 (WebShop) \\
\quad History length & 2 steps \\
\midrule
\textbf{Parallelization} & \\
\quad Number of workers & 8 (evaluation) \\
\quad Process timeout & 3,600 seconds \\
\quad Synchronization style & Dynamic by explorer \\
\midrule
\textbf{Data} & \\
\quad ALFWorld test sets & test\_unseen.jsonl, test.jsonl, train\_hard.jsonl \\
\quad ScienceWorld test sets & Test split from training data \\
\quad WebShop test sets & Test split from training data \\
\bottomrule
\end{tabular}
\end{table}

\subsection{More experiments results}
\label{app: b2f}

\begin{figure}[h]
    \centering
    \includegraphics[width=1\linewidth]{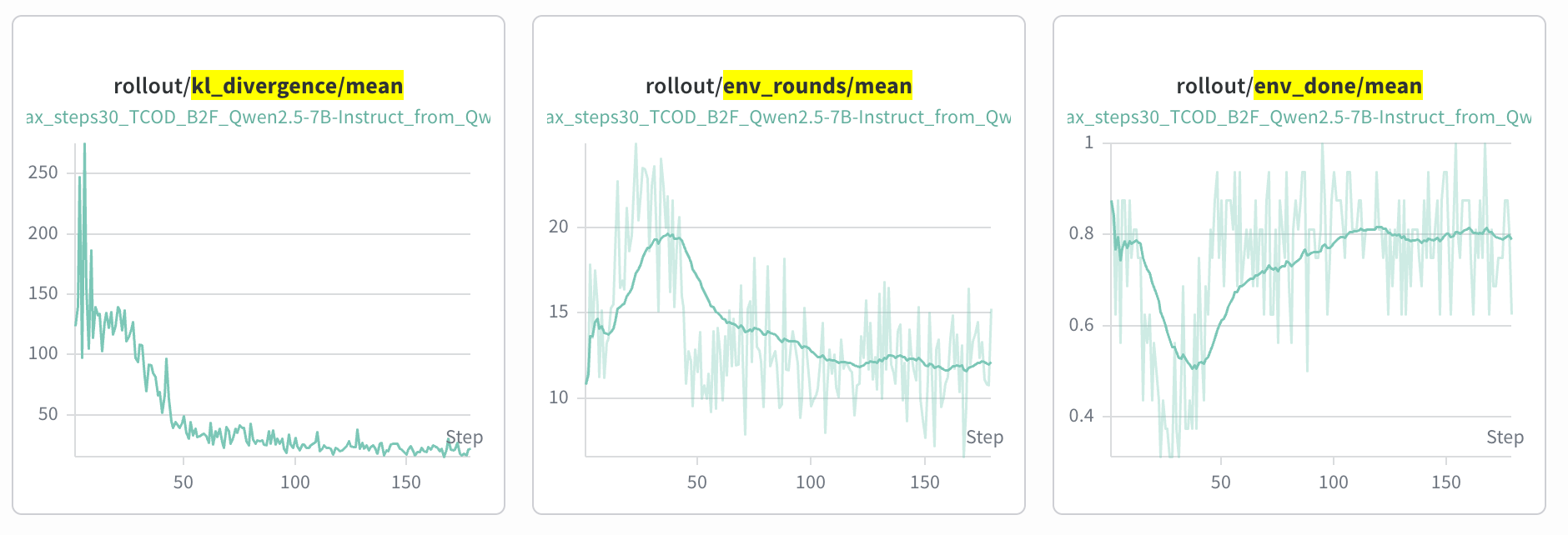}
    \caption{Training dynamics of \ours-B2F ($\eta=2$), including KL divergence, student action horizon, and success rate, for a Qwen2.5-7B student distilled from a GRPO-trained Qwen2.5-7B teacher on ALFWorld.}
    \label{fig:b2f1}
\end{figure}

\begin{figure}[h]
    \centering
    \includegraphics[width=1\linewidth]{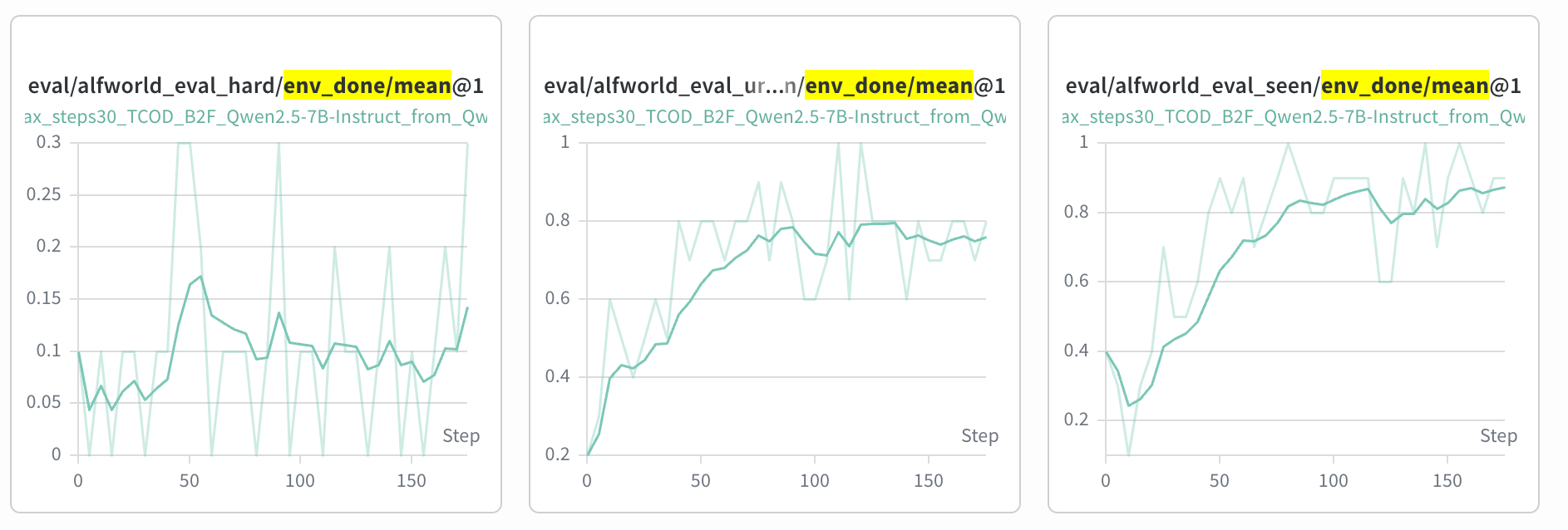}
    \caption{Success rates of \ours-B2F ($\eta=2$), including train hard (left), valid unseen(middle), and valid seen(right), for a Qwen2.5-7B student distilled from a GRPO-trained Qwen2.5-7B teacher on ALFWorld.}
    \label{fig:b2f2}
\end{figure}

\paragraph{Detailed success rate for \ours-B2F} As shown in Figure~\ref{fig:b2f1} and Figure~\ref{fig:b2f2}, \ours-B2F exhibits a characteristic non-monotonic training dynamic. Specifically, the rollout success rate is initially high—since training starts from short horizons—then drops as the curriculum expands to longer trajectories, and finally recovers as the student adapts to the increased difficulty. A similar pattern is observed in the \textit{valid seen} split, where the success rate also decreases mid-training before improving again.

In contrast, the \textit{valid unseen} and \textit{train hard} splits remain relatively stable throughout training, without pronounced drops. This suggests that the intermediate degradation is not due to overfitting or instability, but rather reflects a controlled curriculum transition. Overall, these results indicate that \ours-B2F introduces temporary difficulty as the horizon expands, yet maintains stable generalization while ultimately improving performance, validating the effectiveness of progressive horizon expansion.

\section{Environment Prompts}
\label{appendix:prompts}

This section provides the detailed prompts used for each environment during training and evaluation. All prompts follow a consistent structure: task description, observation-action history, current observation, admissible actions, and thinking/action format requirements.

\subsection{ALFWorld Prompts}

ALFWorld is an embodied AI task requiring agents to navigate household environments and complete object manipulation tasks. The prompt structure emphasizes step-by-step reasoning within \texttt{<thought>} tags followed by executable actions in \texttt{<action>} tags.

\begin{tcolorbox}[
  colback=white,
  colframe=darkblue,
  title=ALFWorld Task Prompt Template,
  fonttitle=\bfseries\small,
  boxrule=0.5pt
]
\small
\texttt{You are an expert agent operating in the ALFRED Embodied Environment. Your task is to: \{task\_description\}}\\
\texttt{Prior to this step, you have already taken \{step\_count\} step(s). Below are the most recent \{history\_length\} observations and the corresponding actions you took: \{action\_history\}}\\
\texttt{You are now at step \{current\_step\} and your current observation is: \{current\_observation\}}\\
\texttt{Your admissible actions of the current situation are: [\{admissible\_actions\}].}

\medskip
\texttt{Now it's your turn to take an action.}\\
\texttt{You should first reason step-by-step about the current situation. This reasoning process MUST be enclosed within <thought> tags.}\\
\texttt{Once you've finished your reasoning, you should choose an admissible action for current step and present it within <action> </action> tags.}
\end{tcolorbox}

\subsection{ScienceWorld Prompts}

ScienceWorld focuses on scientific reasoning tasks in a text-based laboratory environment. The prompt structure guides agents through multi-step experiments requiring domain knowledge and procedural reasoning.

\begin{tcolorbox}[
  colback=white,
  colframe=darkblue,
  title=ScienceWorld Task Prompt Template,
  fonttitle=\bfseries\small,
  boxrule=0.5pt
]
\small
\texttt{Your ScienceWorld task is: \{task\_description\}}\\
\texttt{Prior to this step, you have already taken \{step\_count\} step(s). Below are the most recent \{history\_length\} observations and the corresponding actions you took: \{action\_history\}}\\
\texttt{You are now at step \{current\_step\} and your current observation is: \{current\_observation\}}\\
\texttt{Your valid actions of the current situation are: [\{admissible\_actions\}].}

\medskip
\texttt{Now it's your turn to take an action.}\\
\texttt{You should first reason step-by-step about the current situation. This reasoning process MUST be enclosed within <thought> tags.}\\
\texttt{Once you've finished your reasoning, you should choose a valid action for the current step and present it within <action> </action> tags.}
\end{tcolorbox}

\subsection{WebShop Prompts}

WebShop presents e-commerce shopping tasks requiring agents to navigate product listings, apply filters, and make purchasing decisions based on natural language instructions. The prompt emphasizes matching user preferences to available product attributes.

\begin{tcolorbox}[
  colback=white,
  colframe=darkblue,
  title=WebShop Task Prompt Template,
  fonttitle=\bfseries\small,
  boxrule=0.5pt
]
\small
\texttt{You are an expert autonomous agent operating in the WebShop e-commerce environment.}\\
\texttt{Your task is to: \{task\_description\}.}\\
\texttt{Prior to this step, you have already taken \{step\_count\} step(s). Below are the most recent \{history\_length\} observations and the corresponding actions you took: \{action\_history\}}\\
\texttt{You are now at step \{current\_step\} and your current observation is: \{current\_observation\}.}\\
\texttt{Your admissible actions of the current situation are:}\\
\texttt{[}\\
\texttt{\{available\_actions\}}\\
\texttt{].}

\medskip
\texttt{Now it's your turn to take one action for the current step.}\\
\texttt{You should first reason step-by-step about the current situation, then think carefully which admissible action best advances the shopping goal. This reasoning process MUST be enclosed within <thought> tags.}\\
\texttt{Once you've finished your reasoning, you should choose an admissible action for current step and present it within <action> </action> tags.}
\end{tcolorbox}

\paragraph{Action Format for WebShop.} WebShop uses a specific action format with two primary action types:
\begin{itemize}
    \item \texttt{search[<query>]}: Search for products using a text query (only available when search bar is present)
    \item \texttt{click[<button\_name>]}: Click on interactive elements (e.g., product links, filter buttons, pagination)
\end{itemize}

The available actions are dynamically presented based on the current page state, including clickable elements and search bar availability.



\end{document}